\documentclass{article}

\usepackage{arxiv}

\usepackage[full]{textcomp}

\usepackage{mathdesign}
\usepackage{comment}
\usepackage{multirow}
\usepackage{booktabs}
\usepackage{subcaption}
\usepackage{caption}
\usepackage[dvipsnames]{xcolor} 
\usepackage{pgfplots}
\usepackage{soul}
\usetikzlibrary{calc}

\usepackage[normalem]{ulem}

\usepackage{makecell, cellspace, caption}

\usepackage{amssymb} 
\usepackage{amsfonts}
\usepackage{amsmath}
\usepackage{amsthm}
\usepackage{amsmath,amsfonts,amsthm,bm} 
\usepackage{hyperref} 
\hypersetup{colorlinks}
\usepackage{algorithm} 
\usepackage{algpseudocode} 
\usepackage[switch,displaymath]{lineno}
\usepackage{graphicx}
\usepackage{times}
\usepackage{epsfig}
\usepackage{tabularx}
\usepackage{mathtools}
\usepackage{color, colortbl}
\usepackage{enumitem}
\usepackage{hhline}
\usepackage{bbm} 
\usepackage{array}
\usepackage{pifont}
\usepackage{array, makecell} %
\usepackage{diagbox}
\usepackage{verbatim}
\newcolumntype{C}[1]{>{\centering\let\newline\\\arraybackslash\hspace{0pt}}m{#1}}

\hypersetup{
    colorlinks=true,
    linkcolor=blue,
    filecolor=magenta,      
    urlcolor=cyan,
    pdftitle={Overleaf Example},
    pdfpagemode=FullScreen,
    }

\title{Harmonizing Flows: Leveraging normalizing flows for unsupervised and source-free MRI harmonization}

\newcommand{\xmark}{\ding{55}}
\newcommand{\cmark}{\ding{51}}

\def\gX{{\mathcal{X}}}
\def\gS{{\mathcal{S}}}
\def\gT{{\mathcal{T}}}
\def\gL{{\mathcal{L}}}
\def\gA{{\mathcal{A}}}

\def\gD{{\mathcal{D}}}
\def\gF{{\mathcal{F}}}
\def\gU{{\mathcal{U}}}

\def\vx{{\mathbf{x}}}
\def\vu{{\mathbf{u}}}
\def\vz{{\mathbf{z}}}
\def\vy{{\mathbf{y}}}

\def\sR{{\mathbb{R}}}

\newcommand{\mypar}[1]{\vspace{0.25em}\noindent\textbf{#1}~}
\newcommand{\ppm}{\,\scriptsize$\pm$}
\newcommand{\argmin}{\operatornamewithlimits{argmin}}

\algnewcommand\algorithmicinput{\textbf{Input:}}
\algnewcommand\Input{\item[\algorithmicinput]}
\algnewcommand\algorithmicoutput{\textbf{Output:}}
\algnewcommand\Output{\item[\algorithmicoutput]}

\author{Farzad Beizaee$^{1,2,3,*}$ \And Gregory A. Lodygensky$^{3,4}$ \And Chris L. Adamson$^{5}$ \And Deanne K. Thompson$^{5,6,7}$ \And Jeanie L.Y. Cheong$^{5,7,8,9}$ \And Alicia J. Spittle$^{5,8,10}$ \And Peter J. Anderson$^{5,6}$ \And Christian Desrosiers$^{1,2}$ \And Jose Dolz$^{1,2}$}

\date{$[1]$ LIVIA, ÉTS, Montreal, Canada \\ $[2]$ ILLS, McGill - ETS - Mila - CNRS - Université Paris-Saclay - CentraleSupelec \\ $[3]$ CHU Sainte-Justine, University of Montreal, Canada \\ $[4]$ Canadian Neonatal Brain Platform, Montreal, Canada \\ $[5]$ Murdoch Children’s Research Institute, Victoria, Australia \\ $[6]$ School of Psychological Sciences, Monash University, Australia \\ $[7]$ Department of Paediatrics, The University of Melbourne, Australia \\ $[8]$ The Royal Women’s Hospital, Victoria, Australia \\ $[9]$ Department of Obstetrics and Gynaecology, The University of Melbourne, Australia \\ $[10]$ Department of Physiotherapy, The University of Melbourne, Australia \\ \texttt{*farzad.beizaee.1@ens.etsmtl.ca}}

\renewcommand{\shorttitle}{Harmonizing Flows}

\begin{document}
\maketitle

\begin{abstract}

Lack of standardization and various intrinsic parameters for magnetic resonance (MR) image acquisition results in heterogeneous images across different sites and devices, which adversely affects the generalization of deep neural networks. To alleviate this issue, this work proposes a novel unsupervised harmonization framework that leverages normalizing flows to align MR images, thereby emulating the distribution of a source domain. The proposed strategy comprises three key steps. Initially, a normalizing flow network is trained to capture the distribution characteristics of the source domain. Then, we train a shallow harmonizer network to reconstruct images from the source domain via their augmented counterparts. Finally, during inference, the harmonizer network is updated to ensure that the output images conform to the learned source domain distribution, as modeled by the normalizing flow network. Our approach, which is unsupervised, source-free, and task-agnostic is assessed in the context of both adults and neonatal cross-domain brain MRI segmentation, as well as neonatal brain age estimation, demonstrating its generalizability across tasks and population demographics. The results underscore its superior performance compared to existing methodologies. The code is available at \href{https://github.com/farzad-bz/Harmonizing-Flows}{https://github.com/farzad-bz/Harmonizing-Flows}

\end{abstract}

\setlength{\parskip}{3pt}

\section{Introduction}
\label{sec:intro}

Magnetic Resonance Imaging (MRI) serves as an indispensable tool in modern medical diagnostics, enabling clinicians to obtain detailed insights into anatomical structures and pathological conditions. However, the inherent variability in MRI data acquisition protocols across different imaging sites poses significant challenges in achieving consistent and reliable image interpretation. This variability can stem from differences in scanner hardware, imaging parameters, and patient populations~\cite{takao2011effect}, leading to inconsistencies in image appearance and potentially confounding downstream analysis. For instance, MRIs acquired from two different scanners or with different sets of parameters and configurations will often have noticeable appearance differences, which can be considered as domain shift. Therefore, pooling multi-centric clinic trials to address specific questions does not necessarily enhance statistical power, as the introduced variance may partially stem from non-clinical sources. 

On the other hand, despite the considerable progress observed in deep learning, these models still face challenges in coping with distributional shifts. The performance of deep neural networks in fundamental visual problems, such as classification, segmentation, and regression, largely degrades when they are applied to data acquired under varied conditions, consequently limiting their broad applicability. Specifically, models trained on data from a specific site often struggle to achieve similar performance when applied to images from other centers.

To mitigate this challenge, image harmonization tackles the problem of distributional drifts by mapping images from one domain to another, aiming at transferring contrast characteristics across diverse datasets. MRI harmonization ensures the comparability of MRI data collected from different scanners, facilitating accurate and consistent analysis for multi-center studies involving diverse imaging datasets. 
However, many existing harmonization approaches rely on assumptions that could impede their feasibility and scalability in real-world applications. For example, some methods involve acquiring imaging data of the same anatomical targets from multiple sites or locations. These methods, often referred to as using \textit{traveling subjects}, aim to identify and quantify the transformations needed to harmonize the data across different acquisitions settings~\cite{dewey2019deepharmony,durrer2023diffusion}. Another family of approaches requires access to the source images during harmonization or knowing the target domain in advance ~\cite{pomponio2020harmonization, modanwal2020mri, liu2021style, cackowski2023imunity}, which might not be feasible in practical scenarios. Furthermore, several of these harmonization strategies require annotated data associated with the downstream task~\cite{delisle2021realistic, dinsdale2021deep, karani2021test}. This poses an additional challenge to the harmonization, as acquiring labeled data can be resource-intensive and time-consuming, especially when dealing with large datasets and dense tasks such as segmentation. Finally,  many harmonization techniques require knowledge of the target domains during the training phase, despite the common occurrence of unknown target domains in real-world scenarios. Based on the limitations exposed above, we present the following contributions in this work:
\begin{itemize}
    \item We alleviate the aforementioned constraints on MR harmonization and introduce a novel harmonization approach that is \textit{unsupervised}, \textit{source-free} ($\gS\gF$), \textit{task-agnostic} ($\gT\gA$) and can cope with \textit{unknown-domains} ($\gU\gD$) without necessitating retraining for every target distribution. In fact, our approach only requires MRIs from one modality of the source domain during training, as opposed to existing approaches. 
    \item Specifically, we propose to leverage a modern family of generative models, known as normalizing flows, which have proven to be highly effective in modeling data distributions for generative purposes.  
    
    \item  Alongside the methodological novelty of the proposed method, our empirical findings illustrate that it yields significant improvements over existing harmonization techniques, while effectively mitigating their limitations. More importantly, the comprehensive experimental section on multiple tasks and datasets demonstrates that the proposed approach successfully generalizes across target tasks and population demographics.

\end{itemize}

A preliminary conference version of this work has been presented at IPMI 2023~\cite{beizaee2023harmonizing}. This manuscript provides a substantial extension of the conference version, which includes
\textit{i)} an extended literature review on methods addressing the problem of distribution shift, and a comprehensive empirical validation of the proposed approach, including \textit{ii)} additional recent approaches for harmonization, \textit{iii)} extensive ablation studies to validate our choices, \textit{iv)} assessing the performance of our approach in multiple tasks and population demographics, \textit{v)} including additional adult datasets in the experiments, \textit{vi)} employing additional evaluation metrics to assess the performance from a harmonization standpoint, \textit{vii)} and complimentary plots and results to better understand the overall performance of the different studied methods.   


\begin{figure*}[t!]
\centering
\includegraphics[width=\linewidth]{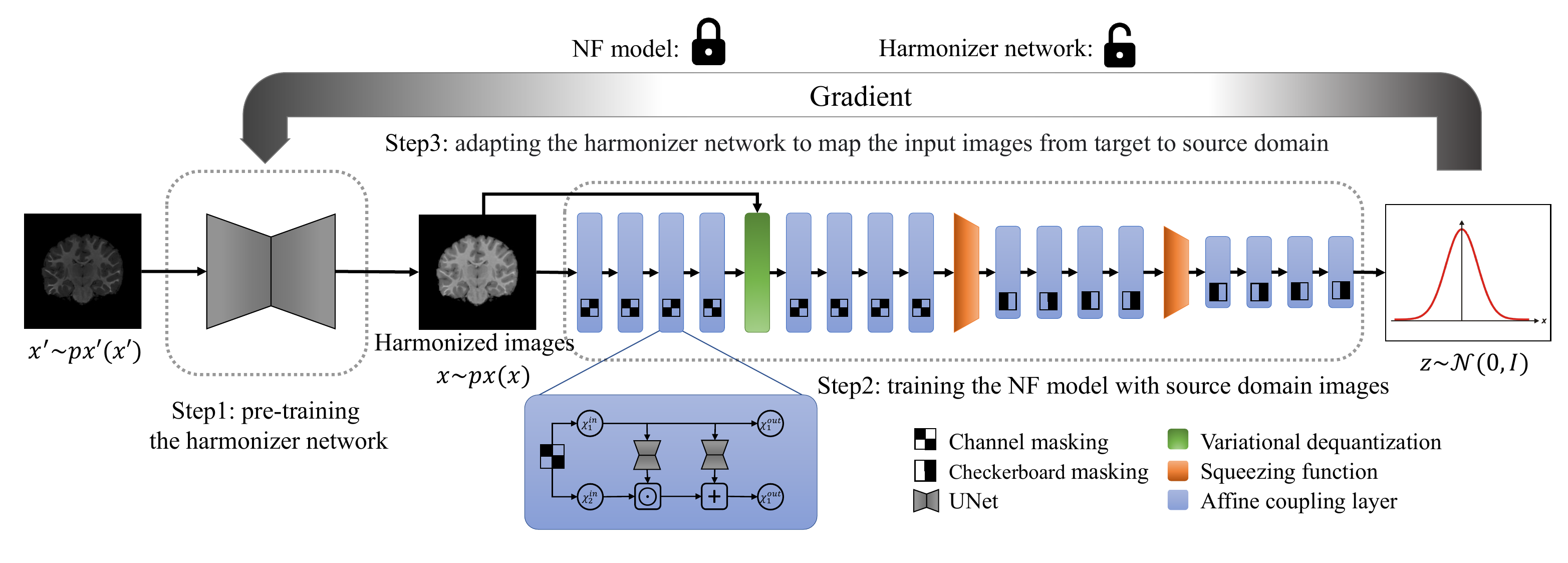}
     \caption{\textbf{Harmonizing Flows Pipeline.} Our method comprises three primary steps. Initially, a harmonizer network undergoes pre-training to reconstruct the original images from augmented counterparts, facilitating initial harmonization. Subsequently, normalizing flow (NF) is utilized to capture the distribution of the source domain. In the third stage (test time), the trained NF is leveraged to update the parameters of the harmonizer network, ensuring maximal alignment between the harmonized outputs and the learned NF distribution. Notably, steps 1 and 2 are independent of each other and can be executed interchangeably.
     }
\label{fig:normalizing flow architecture.}
\end{figure*}

\section{Related work}
\label{sec:related}

\mypar{Image harmonization.}
In the medical domain, various methods have been proposed to harmonize images, with a particular focus on MRI data. Traditional post-processing procedures like intensity histogram matching~\cite{nyul2000new,shinohara2014statistical} help mitigate biases across scanners, but may also eliminate informative local intensity variations. Statistical harmonization approaches, on the other hand, can model both image intensity and dataset bias at the voxel level~\cite{fortin2016removing,fortin2017harmonization,beer2020longitudinal}. However, when the variations in data distribution are more complex and localized, they typically lead to sub-optimal harmonization outcomes. Additionally, these methods must often be adjusted each time images from new sites are provided, further compromising their performance in real-world applications. Modern strategies for image harmonization, using deep learning techniques, hold significant promise as an alternative solution for this problem~\cite{dewey2019deepharmony,zhu2017unpaired,liu2021style,zuo2021information,delisle2021realistic,dinsdale2021deep}. Yet, these approaches often rely on unrealistic assumptions, which pose significant barriers to scalability when applied to extensive multi-site harmonization endeavors. First, some methods require images of the same target anatomy across different sites, known as \textit{traveling subjects}, to identify intensity transformations among different sites~\cite{dewey2019deepharmony,durrer2023diffusion}. This means that a given number of subjects being scanned at every site or scanner is employed for training, a condition rarely met in practice. The most widely used models for MR harmonization are GANs and autoecoders, which have shown promising results in reducing multi-site variation through image-to-image synthesis. GANs perform domain translation by learning domain invariant features. One issue of such models is that they are limited to mapping between two specific scanners for most studies. Also, target domains are required to be known at the training time~\cite{zhu2017unpaired,liu2021style} which is a limiting factor for the scalability of the harmonization process. Additionally, each time a new domain is added, these approaches must be fine-tuned to accommodate the characteristics of this domain. Autoencoder-based methods~\cite{torbati2021multi, dewey2020disentangled, zuo2021information, zuo2023haca3, wu2023structural, cackowski2023imunity}, on the other hand, aim to harmonize data in terms of disentangled representations. This group of harmonization methods attempted to extract scanner-related features for harmonization. Similar to GANs, data from multiple sources and target domains are required for training. CALAMITI~\cite{zuo2021information}, and HACA3~\cite{zuo2023haca3}, which are two key methods of this category, require paired multi-modal MRIs for training, thereby restricting their applicability, particularly in single-modality scenarios. DLEST~\cite{wu2023structural} and Imunity~\cite{cackowski2023imunity} try to solve the problem of requiring multi-modal data for latent disentanglement. However, training these methods can be challenging due to the instability of their adversarial learning strategy. Lastly, task-dependent methods leverage labels associated with each image for a particular downstream task to optimize the harmonization for this specific problem~\cite{delisle2021realistic,dinsdale2021deep}. These methods often rely on task-specific features or assumptions, making them less effective when applied to new tasks or unseen data. Additionally, task-dependent harmonization approaches require large amounts of annotated data for training, which can be costly and time-consuming to acquire.

\mypar{Test-time Domain Adaptation.}  Traditional solutions to the problem of distributional shift use labeled samples from a source domain and unlabeled ones from the target domain for adapting a source-trained model to perform well on the target. Several strategies for this task, known as unsupervised domain adaptation (UDA), work by explicitly aligning the feature distributions of the source and target domains~\cite{wang2023shape,wu2020cf}. Another popular approach consists in learning a domain-agnostic representation, for example using adversarial networks~\cite{dou2019pnp,kamnitsas2017unsupervised}. Generative adversarial networks (GANs) can also be employed in style transfer methods to change the appearance of images from the target to the source while also preserving their semantic structures~\cite{chen2020unsupervised,zhao2021mt}.

A major limitation of UDA methods is their need to have source examples during the adaptation phase, which may be impracticable in medical applications due to data sharing restrictions. Source-free domain adaptation (SFDA) approaches~\cite{bateson2022source,yang2022source,stan2024unsupervised} relax this constraint and instead adapt the source-trained model using only unlabeled data from the target domain. However, these approaches are usually task-dependent and require an explicit adaptation process for each new target domain, hence are not compatible with the harmonization setting investigated in this work.

Another strategy for addressing distribution shift, more closely related to our method, is test-time adaptation (TTA)~\cite{boudiaf2022parameter,mummadi2021test,liang2020we,wang2021tent,niu2022efficient,niu2023towards}. Unlike SFDA, which performs adaptation in an offline step, this strategy adapts a pre-trained deep neural network to domain shifts encountered \emph{during inference} on test samples. One of the earliest TTA approaches, called TENT~\cite{wang2021tent}, updates the normalization layers of the network by minimizing the Shannon entropy of predictions for test samples. In~\cite{mummadi2021test}, entropy minimization has been changed by optimizing a log-likelihood ratio and considering the normalization statistics of the test batch. EATA~\cite{niu2022efficient} instead introduces an active sample selection criterion to identify reliable and non-redundant samples. The model is then updated based on these samples to minimize entropy loss for test-time adaptation. In~\cite{niu2023towards}, authors propose a sharpness-aware and reliable entropy minimization method called SAR, further stabilizing the TTA process. Moreover, SHOT~\cite{liang2020we} adapts the entire feature extractor with a mutual information loss, while using pseudo-labels to provide additional test-time guidance. Instead of updating the network parameters, LAME~\cite{boudiaf2022parameter} uses Laplacian regularization to do a post-hoc adaptation of the softmax predictions. Recent works have explored the potential of TTA for cross-site/modality segmentation of medical images. 
Authors of \cite{hu2021fully} propose a TTA method for segmentation using a  regional nuclear-norm loss to improve the discriminability and diversity of predictions and a contour regularization term to enforce segmentation consistency between nearby pixels. 
\textcolor{black}{Contrary to this paradigm, where typically the target-task network (e.g., classification or segmentation) is adapted at inference based on surrogate losses on the network predictions, our work focuses on modifying the image appearance instead, which offers a more general solution, which is agnostic to the task at hand.}

Furthermore, while \cite{karani2021test} uses the reconstruction error of an auto-encoder applied on segmentation outputs to normalize input images, it requires segmentation masks for the adaptation, which makes of this strategy task and annotation dependent. 

\mypar{Normalizing flows.}
Popular methods for generative tasks include generative adversarial networks (GANs)~\cite{goodfellow2020generative} and Variational Auto-encoders~\cite{kingma2014auto}. Despite their popularity and wide acceptance, these methods present several important limitations, including mode~\cite{salimans2016improved} and posterior collapse~\cite{lucas2019understanding}, training instability~\cite{salimans2016improved}, and the incapability of providing an exact evaluation of the probability density of new data points. 
Recently, normalizing flows (NF) have emerged as a popular approach for constructing probabilistic and generative models due to their ability to model complex distributions~\cite{dinh2016density}. Normalizing flows involves mapping a complex distribution, often unknown or poorly characterized, to a simpler distribution, typically the standard normal distribution. This is accomplished through a series of  invertible and differentiable transformations. These transformations allow for efficient density estimation, sampling, and generative modeling. One of the key advantages of normalizing flows is their ability to capture intricate dependencies within data while providing tractable likelihood estimation, enabling a wide range of applications across domains. While the majority of current literature has utilized NFs for generative purposes (e.g., image generation~\cite{ho2019flow++,kingma2018glow}, noise modeling~\cite{abdelhamed2019noise}, graph modeling~\cite{zang2020moflow}) and anomaly detection~\cite{gudovskiy2022cflow,kirichenko2020normalizing}, recent findings also indicate their effectiveness in aligning a given set of source domains~\cite{grover2020alignflow,usman2020log, osowiechi2023tttflow}. Closely related to our problem, and up to the best of our knowledge, only a few attempts have investigated using normalizing flows to harmonize MR images. In particular, \cite{wang2021harmonization} presented a strategy that harmonizes pre-extracted features, i.e., brain ROI volume measures, and not image harmonization as in this work. In addition, extracting these ROIs requires pixel-wise labels, making of this approach task-dependent, contrary to our method which is task-agnostic. Furthermore, \cite{jeong2023blindharmony} is a concurrent approach that appeared after the conference version of this work was published. As discussed in their work, and shown empirically in our evaluation, although this method also leverages normalizing flows to harmonize images, it fails in the presence of large domain drifts between source and target sites. 


\section{Methodology}
\label{sec:methodology}

We first define the problem addressed in this study. Consider $\gX_\gS=\{\vx_{n}\}_{n=1}^{N}$ as a collection of unlabeled images from the source domain $\mathcal{S}$, where each image $i$ is represented by $\vx_i \in \sR^{|\mathrm{\Omega}|}$, with $\mathrm{\Omega}$ indicating its spatial domain (i.e., $W\!\times\!H$). Likewise, let $\gX_\gT=\{\vx_{n}\}_{n=1}^{M}$ be the set of unlabeled images within a target domain $\gT$\,\footnote{
For simplicity, we assume a single domain exists here. However, our formulation can be readily extended to accommodate $T$ distinct domains.}. The objective of unsupervised data harmonization is to discover a mapping function $f_{\theta} : \gT \!\rightarrow\!\gS$ without relying on labeled images or paired data from either domain. 

We introduce a solution based on normalizing flows to address this problem. The proposed framework, which comprises three separate steps, is illustrated in \autoref{fig:normalizing flow architecture.}. In Step 1, we first utilize Normalizing Flows, renowned for their ability to accurately learn data likelihoods, to capture the source domain's distribution. In Step 2, we then employ an auto-encoder as a harmonizer network, pre-training it by reconstructing original images from the augmented source domain counterparts. Finally, during testing, we update the parameters of the harmonizer network using images from the unseen target domain, ensuring that the harmonized outputs align with the learned distribution using NF model. The following sections present each of these steps in greater detail.

\setcounter{footnote}{0} 

\subsection{Learning the source domain distribution}
\label{ssec:NFs}

We have leveraged normalizing flows~\cite{dinh2016density} to model the source domain distribution. NFs are a modern family of generative models capable of modeling complex probability density $p_{x}(\vx)$ (i.e., the source domain distribution) through applying a sequence of transformation functions, denoted as $g_{\phi}=g_1 \circ g_2 \circ \dots g_T$, on known simple probability density $p_{u}(\vu)$ such as standard normal distribution. Source image can be represented as $\vx=g_{\phi}(\vu)$, where $\vu \sim p_{u}(\vu)$ and $p_{u}(\vu)$ denotes the base distribution of the normalizing flow model. An essential condition for the transformation function $g_{\phi}$ is its requirement to be \textit{invertible}, with both $g_{\phi}$ and $g_{\phi}^{-1}$ being \textit{differentiable}. With these prerequisites met, the density of the original variable $\vx$  is well-defined, allowing for the exact computation of its likelihood using the change of variables rule, expressed as:
\begin{equation}
\begin{aligned}
 \log  p_{\mathrm{x}}(\vx) \, &= \, \log p_{\mathrm{z}}\left(g^{-1}_{\phi}(\vx)\right) + \log\left|\operatorname{det} \big(\mathbf{J}_{g^{-1}_{\phi}}(\vx)\big)\right| \\
 &= \log p_{\mathrm{z}}\left(g^{-1}_{\phi}(\vx)\right) + \sum_{t=1}^T \log\left|\operatorname{det} \big(\mathbf{J}_{g^{-1}_{t}}(\vu_{t-1})\big)\right|
\end{aligned}
\label{eq:NF}
\end{equation}
The first component of the right side corresponds to the log-likelihood within the simple distribution and $\mathbf{J}_{g_t^{-1}}(\vu_{t-1})$ indicates the Jacobian matrix corresponding to the transformation $g_t$. In order to train the Normalizing flow model and learn the source domain distribution, model parameters $\phi$ are learned by maximizing the likelihood of the transformed data under the simpler distribution. This is achieved by minimizing the negative log-likelihood in Eq. \ref{eq:NF} which leads to the following loss function:
\begin{equation}
\gL_{NF} \, = \, -\log  p_{\mathrm{x}}(\vx)
\label{eq:NF_Loss}
\end{equation}

\mypar{Building the Normalizing Flow.}
Constructing a bijective transformation function neural network for the Normalizing Flow (NF) model often involves the stacking of affine coupling layers, as highlighted by~\cite{dinh2016density,kingma2018glow}. This approach has been established as an efficient strategy. Coupling layers offer computational symmetry, meaning they are equally rapid in both evaluation and inversion processes. This characteristic addresses usability concerns inherent in asymmetric flows, such as masked autoregressive flows, which makes the coupling layers a preferred choice. Their balanced computational efficiency enables smoother integration into various applications, contributing to their widespread adoption in NF architectures. Suppose $\vz \in \sR^D$ serves as the input to the coupling layer, which is partitioned disjointly into ($\vz^A, \vz^B) \in \sR^d \times \sR^{D-d}$. The partitioning can be done along spatial dimensions(e.g. checkerboard masking strategy) or channels (channel masking strategy). Then, the transformation function $g(\cdot):\sR^D \rightarrow \sR^D$ can be expressed as:
\begin{equation}
\vy^A =\vz^A, \quad \vy^B =\vz^B \odot \exp \left(s\left(\vz^A\right)\right)+t\left(\vz^A\right)
\label{eq:coupl}
\end{equation}
where $\vy^A$ and $\vy^B$ represent the transformed parts of the input data and $\odot$ is element-wise multiplication. This formulation divides the input into two parts ($\vz^A$ and $\vz^B$), and transforms only the latter part, leaving the former unchanged. This setting offers simplicity for calculating the Jacobian determinant, which makes it possible to use complex neural networks as shift $s(\cdot)$ and scale $t(\cdot)$ networks. Note that the transformation in Eq. \ref{eq:coupl} is invertible and therefore allows for efficient recovery of $\vz^A$ and $\vz^B$ from $\vy^A$ and $\vy^B$. The work in~\cite{dinh2016density} presented coupling flows on simpler tasks and datasets which demanded less complex representations. However, our current task necessitates pixel-to-pixel mappings on more challenging data. Therefore, we replace the simple convolutional blocks in~\cite{dinh2016density} with shallow U-shaped convolutional neural networks to find the scale and shift parameters of the affine transformation, as they capture broader contextual information and provide higher representational capacity. 
Moreover, given that NFs rely on the change of variables rule, which operates within continuous space, it is important to ensure that the input is continuous. Traditionally, dequantization involves adding uniform noise $u\!\in\!U[0,1]$ to discrete values to convert them into continuous representations. However,  it could lead to a hypercube representation with sharp borders. Such sharp borders pose a challenge for modeling with a flow, as it relies on smooth transformations. Recently, a variational framework was introduced~\cite{ho2019flow++} to expand dequantization to more sophisticated distributions. This was achieved by substituting the uniform distribution with a learnable distribution. This learnable distribution can be optimized alongside other parameters of the normalizing flow, allowing for end-to-end training and seamless integration into the density estimation process.

\mypar{Constraining the source-distribution learning.} Optimizing the objective in Eq. \ref{eq:NF_Loss} solely with source images could potentially bias the model towards emphasizing characteristics of subjects, such as age and gender, rather than focusing on source-specific attributes like contrast and brightness. To address this concern, we propose a strategy aimed at facilitating the learning of the source domain distribution. For this purpose, in each iteration, we randomly select $N'$ images of the source dataset $\gX_{\gS}$ and apply a series of augmentations $f_{aug}(\cdot)$ in such a way that the resulting image exhibits a dissimilarity in appearance compared to the original image, as measured by the mean squared distance, surpassing a predetermined threshold. These images can be served as out-of-distribution samples, to guide the normalizing flow model to learn source-specific characteristics.
In particular, we employ different types of contrast augmentation, brightness changes, multiplicative transformations, and random monotonically increasing mapping functions to augment these images. Then, the overall learning objective of our model can be defined as follows.
\begin{equation}
\label{eq:overall}
\gL_T = 
\underbrace{-\sum_{n=1}^{N-N'} \log  p_{\mathrm{x}}(\vx_n)}_{\textrm{Source distribution modeling}} - \underbrace{\sum_{n=1}^{N'} \min\left(\mathrm{c}, -\log  p_{\mathrm{x}}(f_{aug}(\vx_n))\right)}_{\textrm{Guiding term}}.
\end{equation}
The first term is the learning objective in Eq. \ref{eq:NF_Loss} over the source images, while the second one encourages the NF model to reduce the likelihood of the augmented images, which facilitates the learning of domain-specific characteristics rather than subject-related features. Furthermore, to prevent divergence of the negative log-likelihood for an augmented sample  to infinity, we employ a constant margin in the second term denoted as $\mathrm{c}$.

\subsection{Achieving image harmonization}
\label{ssec:harmo}

\mypar{Harmonizer network.} The goal of our harmonizer network $h_{\theta}(\cdot)$ is to perform image-to-image translation of MRIs from the target to the source domain in such a way that:
$p_{\mathbf{x}}(\mathbf{x})=p_{\mathbf{x}'}(h_{\theta}(\mathbf{x}'))$. It's important to highlight that the input and output of the harmonizer network must share the same spatial dimensions. Here $\theta$ denotes the set of learnable parameters within the harmonizer network, and $\vx$ and $\vx'$ represent the samples from the source domain and target domains, respectively. This implies that the harmonizer network aims to shift the distribution of the target images so that they align with the distribution of source images. Nonetheless, we want the proposed method to operate effectively on unseen domains, necessitating that the target domains remain unknown during training. To this end, at first, the harmonizer network was trained to restore the original MRIs of the source domain from its augmented version. As in the previous step, we used different types of contrast augmentation, brightness changes, multiplicative transformations, and random monotonically increasing mapping functions. In contrast to the first step, there are no restrictions on how much the image can be altered as long as the augmented image is logical and details are not eliminated. To train this model, we employed the sum of two commonly used standard reconstruction loss functions: SSIM (Structural Similarity Index Measure) loss~\cite{wang2004image} and L1 loss (mean absolute error). SSIM loss is more appropriate when preserving structural similarity and perceptual quality is important, while L1 loss is suitable for tasks where exact pixel-wise accuracy is required, regardless of perceptual differences.  The learning objective for the harmonizer network thus becomes:
\begin{equation}
\begin{split}
{\theta^{init}} \, = \, \argmin _{\theta} \frac{1}{N} (\sum_{n=1}^{N} \left\|(\vx_{n} - h_{\theta}\left(f_{aug}(\vx_{n})\right)\right\| +\\  SSIM(\vx_{n}, h_{\theta}\left(f_{aug}(\vx_{n})\right)))
\end{split}
\label{eq:harmo}    
\end{equation}

A simple UNet has been considered as the harmonizer network. Also, it's important to emphasize that the conducted augmentations may not perfectly represent potential unseen target domains. So, directly using the trained parameters $\theta^{init}$ for image-to-image translation yields sub-optimal harmonization. Nevertheless, it provides a good starting point for the next phase.

\mypar{Adapting the harmonizer network leveraging the Normalizing Flow.} So far, we have obtained an initial harmonizer network that gives us the possibility of transforming image appearance across domains, while bearing in mind that the output is sub-optimal. Also, we have learned the exact distribution of the source domain using a normalizing flow model, where we ensured that it focuses on domain-specific characteristics. The final step involves refining the harmonizer network to effectively map images from the target domain onto the distribution of the source domain. For this purpose, firstly, we stack the trained NF model at the top of the pre-trained harmonizer network as the \autoref{fig:normalizing flow architecture.}. Note that as we aim to leverage the learned distribution by the NF model which is already  trained with the source data, its parameters should remain frozen during the adaptation of the harmonizer. Then we try to optimize the parameters of the harmonizer network, such that harmonizer network outputs exhibit a high likelihood of aligning with the source domain distribution under the supervision of the trained NF model. The learning objective of the adaptation stage is to increase the likelihood of the harmonizer outputs for images from the target domain, based on the the density estimation provided by the NF model. This objective is encapsulated in a defined loss function, which can be expressed as follows:
\begin{equation}
\gL_{Adap} = 
-\sum_{m=1}^{M}\log  p_{\mathrm{x}}\big(g_{\phi}\big(h_{\theta}(\mathbf{x}_{m})\big)\big) 
\label{eq:adap_loss}    
\end{equation}

As a stopping condition for updating the harmonizer, we assess two potential alternatives: One criterion involves assessing the Shannon entropy of the predictions generated by the target downstream task using the harmonized images (e.g., classification or segmentation), and stopping the adaptation when the entropy plateaus. Additionally, we take into account the bits per dimension (\textit{BPD}), which is a scaled variation of the \textit{negative log-likelihood} commonly utilized for assessing generative models:
\begin{equation}
\textit{BPD}=- \log  p_{\mathrm{x}}(\vx) \cdot\Big(\log 2 \cdot \prod_i \mathrm{\Omega}_i\Big)^{-1}
\label{eq:BPD}    
\end{equation}
where $\mathrm{\Omega}_1$, ..., $\mathrm{\Omega}_T$, are the spatial dimensions of the input images. More concretely, we can stop updating the harmonizer network when the reached \textit{BPD} value matches the BPD observed for the source images using the trained NF model. In practice, the source BPD value can be determined during the training time using a validation set.

\begin{table*}[h!]
\footnotesize
\setlength{\tabcolsep}{3pt}
\caption{\label{tab:Dataset-sites} \textbf{Acquisition parameters across different sites.} Scanner details and phenotypic information for each site used in this study. \emph{y}: years; \emph{gw}: gestation weeks.}
\makebox[\textwidth]{\begin{tabular}{l|c|c|c|c|c|c|c}
\toprule
Sites & \# used MRIs & Scanner & TR (ms) &TE (ms) & Flip angle & Voxel size (mm$^3$) & Age  \\
\midrule
CALTECH & 19 & Siemens & 1590 & 2.73 & 10 & $1.0\times1.0\times1.0$ & 17--56 y \\
KKI & 20 & Phillips & 8 & 3.70 & 8  & $1.0\times1.0\times1.0$ & \phantom{0}8--13 y\\
NYU & 20 & Siemens & 2530 & 3.25 & 7  & $1.3\times1.0\times1.3$ & \phantom{0}6--39 y\\
PITT & 20 & Siemens & 2100 & 3.93 & 7  & $1.1\times1.1\times1.1$ & \phantom{0}9--35 y\\
HBNSI & 77 & Siemens & 2730 & 1.64 & 7 & $1.0\times1.0\times1.0$ & \phantom{0}5--21 y\\
HBNRU & 57 & Siemens & 2500 & 3.15 & 8 & $0.8\times0.8\times0.8$ & \phantom{0}5--21 y\\
OASIS & 117 & Siemens & 9.7 & 4.0 & 10 & $1.0\times1.0\times1.25$  & 18--96 y \\
\midrule
DHCP T1-w & 280 & Philips & 4795 & 8.7 & N/A & $0.8\times0.8\times0.8$ & 37--45 gw\\
DHCP T2-w & 333 & Philips & 12000 & 156 & N/A & $0.8\times0.8\times0.8$ & 37--45 gw\\
V2LP T1-w & 122 & Siemens & 2100 & 3.39 & 9 & $1.0\times1.0\times1.0$ & 37--45 gw\\ 
V2LP T2-w & 122 & Siemens & 8910 & 152 & 120 & $1.0\times1.0\times1.0$ & 37--45 gw\\

\bottomrule
\end{tabular}}
\end{table*}

\section{Experiments}

\subsection{Experimental setting}
First, we resort to the cross-site brain MRI segmentation task to evaluate the harmonization performance of different methods. We chose this task as it allows us to assess not only how effectively the proposed method aligns target domain images with the source domain, but also to evaluate whether the structural details are preserved well during the harmonization process. 
Furthermore, 
we investigate how well the proposed approach performs across different populations, encompassing both neonates and adults, and imaging modalities (i.e., T1-weighted and T2-weighted MRI). 
Last, to demonstrate the generalizability of our method, we explore its performance on the distinct task of neonatal brain gestational age estimation. 

\mypar{Datasets} 

\noindent \textit{MRI harmonization.} It is important to recall that even though the empirical validation showcases the results across all the available sites, the proposed model only has access to a unique domain, i.e., the source, during the training steps and a unique target domain during the harmonization step. The details of the different datasets used for each task are provided below.

\noindent \textit{Adult brain MRI segmentation.} In the context of adult brain MRI segmentation, we utilized data from a total of seven different sites\footnote{Please note that in the conference version of this work~\cite{beizaee2023harmonizing}, only four datasets were employed in the experiments, which explains the differences in the empirical results from both versions.}. Four of these sites are drawn from the Autism Brain Imaging Data Exchange (ABIDE)~\cite{di2014autism} dataset, which includes: California Institute of Technology (CALTECH), Kennedy Krieger Institute (KKI), University of Pittsburgh School of Medicine (PITT), and NYU Langone Medical Center (NYU) sets. Out of the remaining sites, Staten Island (SI) and Rutgers University (RU) sites are sourced from the Healthy Brain Network (HBN)~\cite{alexander2017open} dataset, which we refer to as HBNSI and HBNRI in this paper, along with data from the Open Access Series of Imaging Studies (OASIS)~\cite{marcus2007open}. 
We selected T1-weighted MRIs of a healthy control population from each site, which were skull-stripped, motion-corrected, and quantized to 256 intensity levels. For each site, 60\% of the images are used as the training set, 15\% as the validation set, and the remaining 25\% for testing, which are exploited in a 2D manner using the coronal plane. Moreover, following other large-scale studies~\cite{dolz20183d}, we used Freesurfer~\cite{fischl2012freesurfer} to obtain the segmentations and grouped them into 15 labels: background, cerebral GM, cerebral WM, cerebellum GM, cerebellum WM, CSF, ventricles, brainstem, thalamus, hippocampus, putamen, caudate, pallidum, amygdala and ventral DC.

\noindent \textit{Neonatal brain MRI segmentation and age estimation.} We employed T1-weighted and T2-weighted MRIs from the developing Human Connectome Project (DHCP)~\cite{makropoulos2018developing}, VIBES2~\cite{spittle2014neurobehaviour} and LaPrem~\cite{cheong2021impact} datasets. As VIBES2 and LaPrem datasets are acquired with the same imaging device and parameters, we have combined them and considered them as one site, which we refer to as V2LP hereafter in this paper. All MRIs were sourced from a healthy control population aged between 37 and 45 weeks of gestational age. For segmentation, we utilized the coronal view slices, while the axial view slices were chosen for brain age estimation due to their richer information content. The preprocessing and data splitting procedures for neonates were consistent with those used for adults, with the only difference being the inclusion of contours of 35 regional structures obtained from M-CRIB-S~\cite{adamson2020parcellation}. Image acquisition parameters and device, number of used MRIs, and population age can be found in  \autoref{tab:Dataset-sites} for both adults and neonatal datasets. These values showcase how the selected datasets have distinct imaging devices and parameters.

\mypar{Harmonization baselines.}
The proposed method is compared to a set of  harmonization and image-to-image translation approaches. First, we apply either the segmentation or age regression network directly on non-harmonized images, which we refer to as a "Baseline", so we can assess the improvement gained using the harmonized images. Furthermore, our comparison also includes the following harmonization strategies: Histogram Matching~\cite{nyul2000new}, Combat~\cite{pomponio2020harmonization}, SSIMH~\cite{guan2022fast}, two popular generative-based approaches, i.e., Cycle-GAN~\cite{modanwal2020mri} and Style-Transfer~\cite{liu2021style}, a source free latent-disentanglement harmonization (Imunity)~\cite{cackowski2023imunity}, and a recent method for harmonization based on normalizing flows (BlindHarmony)~\cite{jeong2023blindharmony}.

\mypar{Test time domain adaptation and generalization baselines.}
In addition to the existing harmonization methods, the proposed approach is benchmarked against several test time domain adaptation and generalization methods. These methods include: aleatoric uncertainty estimation (AUE)~\cite{wang2019aleatoric}, which uses test time augmentation to adapt to the target domain; BigAug~\cite{zhang2020generalizing}, which uses heavy augmentations on MRIs for generalization; and TENT~\cite{wang2021tent} and SAR~\cite{niu2023towards}, which are test-time adaptation methods based on the segmentation's output confidence, i.e., entropy. The comparison with these methods aims to shed light on the importance of MRI harmonization and to reveal whether MRI harmonization can be replaced by domain generalization or test-time adaptation methods.

\begin{table*}[t!]
\caption{ \textbf{Performance overview on the cross-site adult MRI segmentation task.} Segmentation performance, in terms of DSC and HD95 metrics, across different harmonization approaches. To facilitate the strengths and weaknesses of different methods, we also indicate whether they are \textit{source-free} $(\mathcal{SF})$, \textit{task-agnostic} $(\mathcal{TA})$, and can handle \textit{unknown-domains} $(\mathcal{UD})$, as well as the different strategy they fall in. The best results are highlighted in \textbf{bold}.} 
\begin{center}
\footnotesize
{\begin{tabular}{l|c|ccc|c|c}
\toprule

Method & Strategy & $\mathcal{SF}$ & 
 $\mathcal{TA}$ & 
 $\mathcal{UD}$ & DSC (\%) & HD95 (mm)  \\
 \midrule 

Baseline  & -- & -- & -- & -- &  36.3\ppm4.8 & 43.1\ppm5.3 \\

Hist matching~\cite{nyul2000new} & Harmonization & \cmark &  \cmark  & \cmark & 63.7\ppm4.8 & 12.3\ppm2.7\\
Combat~\cite{pomponio2020harmonization} & Harmonization & \xmark &  \cmark & \xmark &  73.0\ppm4.3 & 6.3\ppm2.6 \\
Cycle-GAN ~\cite{modanwal2020mri} & Harmonization & \xmark &  \cmark  & \xmark &  75.0\ppm2.8 & 5.5\ppm1.7  \\
Style-transfer~\cite{liu2021style} & Harmonization & \xmark &  \cmark & \xmark & 70.8\ppm5.7 & 8.2\ppm2.6  \\
SSIMH$_\text{ MLMI'22}$~\cite{guan2022fast} & Harmonization & \xmark &  \cmark  & \cmark & 59.4\ppm5.1 & 12.3\ppm2.9 \\
ImUnity$_\text{ MedIA'23}$~\cite{cackowski2023imunity} & Harmonization & \xmark &  \cmark  & \cmark & 58.2\ppm4.6 & 17.0\ppm3.3 \\
BlindHarmony$_\text{ ICCV'23}$~\cite{jeong2023blindharmony} & Harmonization & \cmark &  \cmark  & \cmark & 62.2\ppm6.1 & 13.7\ppm3.2 \\
AUE ~\cite{wang2019aleatoric} & Test Time Augmentation & \cmark & \xmark & \cmark & 35.4\ppm3.6 & 24.5\ppm3.6 \\
BigAug$_\text{ TMI'20}$~\cite{zhang2020generalizing} & Generalization & \cmark & \xmark & \cmark & 82.0\ppm2.1 & 3.2 \ppm0.8  \\ 
TENT$_\text{ ICLR'21}$~\cite{wang2021tent} & Test Time Adaptation & \cmark &  \xmark & \cmark &  72.8\ppm3.4 & 7.4\ppm2.2 \\
SAR$_\text{ ICLR'23}$~\cite{niu2023towards} & Test Time Adaptation & \cmark &  \xmark & \cmark & 70.9\ppm4.0 & 9.9\ppm2.6 \\
\midrule
\textbf{Harmonizing Flows}  & Harmonization  &  \cmark &  \cmark  & \cmark &  \textbf{83.1\ppm2.2} &  \textbf{3.1\ppm0.9} \\
\bottomrule
\end{tabular}}
\end{center}

\label{tab:Results-comparison-adults}
\end{table*}

\mypar{Evaluation metrics and protocol.}
\noindent \textit{Segmentation.} To evaluate the performance of the proposed MRI harmonization approach on adult cross-site brain MRI segmentation, 
we train a segmentation neural network $S_{\Phi}(\cdot)$ using the training set of the source domain. Then for each target domain, the segmentation performance is evaluated using the harmonized images, where the segmentation performance is measured with the Dice Similarity Coefficient (DSC) and 95th percentile Hausdorff Distance (HD95). We repeat these steps considering each site as a source domain and the remaining sites as the target domains in each iteration, and then we report the average of these metrics. The evaluation on neonatal brain MRI segmentation follows the same protocol as that of the adults, which is repeated for each of the two modalities considered, i.e., T1-weighted and T2-weighted. 


\noindent \textit{Neonatal brain age estimation.} The neonatal brain gestational age estimation task follows the same procedure, except that instead of the segmentation model, a deep regression network has been trained. Following the nature of the task, the metrics used to assess its performance are the Mean Absolute Error (MAE) and the Mean Square error (MSE).

\noindent \textit{Harmonization performance.} Last, following~\cite{parida2024quantitative}, we resort to the Wasserstein distance (WD)~\cite{kantorovich1960mathematical} between the normalized intensity histograms of harmonized images and those of the source domain. 
The WD calculates the smallest ``cost'' to change one distribution into the other, taking into account both the amount of change needed and how far the changes need to be shifted. Thus, the Wasserstein distance between two normalized histograms tells us how different they are by considering not only the differences in terms of change magnitudes (such as the KL divergence does), but also how far apart the differing parts are. 
In our experiments, and as stated earlier, we consider one single site as a source domain (while all the remaining domains/sites remain unknown). Furthermore, we report the average of the WD for all source-domain pairs.


\mypar{Implementation details.}

\noindent \textit{Normalizing Flow (FL).} The NF model has been trained for 20,000 iterations with Adam optimizer and a batch-size of 32, an initial learning-rate of $1\times10^{-3}$, and using a weight-decay of 0.5 every 2000 iterations. 
We employ a U-shaped architecture within the coupling layers, comprising four different scales with a scaling factor of 2. Each layer consists of an activation function followed by a convolutional layer and a normalizing layer. The activation function used in each scale is a modified version of ELU, i.e., concat[ELU($x$), ELU($-x$)], which makes it easier for the NF model to map to normal distribution due to its symmetry properties.  Additionally, there are 16, 32, 48, and 64 kernels in each scale, respectively. To construct the NF model, we initially employ four sequential coupling layers with checkerboard masking, aimed at capturing the noise distribution through variational dequantization. Then, this is followed by four identical coupling layers and a squeezing function as explained in~\cite{dinh2016density} to decrease the spatial dimension. Then, we subsequently incorporate four more coupling layers employing a channel-masking strategy, and an additional layer of feature squeezing, followed by a final series of four coupling layers employing channel-masking. The overall architecture of the flow model is shown in  \autoref{fig:normalizing flow architecture.}. Also, the margin $\mathrm{c}$ used for guiding the NF model (Eq. \ref{eq:overall}) is empirically set to 1.2. For adapting the harmonizer network using images of the target domain during test time, the Normalizing Flow model is frozen, and the harmonizer model is updated slightly using Adam optimizer with a learning rate of $5\times10^{-7}$, and a batch-size of 32 until the epoch where the stopping criterion has been reached. All the models were implemented on PyTorch using two NVIDIA RTX A6000 GPU cards.

\noindent \textit{Harmonizer.} The utilized UNet network as harmonizer consists of five different scales with a scaling factor of 2, where each scale includes a layer of the modified ELU activation function, i.e., concat[ELU($x$), ELU($-x$)], followed by two convolutional layers. The number of kernels of the convolutional layers for each scale is 16, 32, 48, 64, and 64 respectively. 

\noindent \textit{Segmentation and regression models.} The segmentation network employed in our empirical validation is nn-UNet~\cite{isensee2021nnu} with batch-normalization, stride 2, and kernel size of 3, whereas ResNet18~\cite{he2016deep} serves as the age estimation network. Furthermore, all the segmentation, age estimation, and harmonizer networks are trained for 5000 iterations with Adam optimizer with an initial learning rate of $1\times10^{-3}$, a weight decay of 0.5 every 500 iterations, and a batch-size of 64. Last, the model at the best iteration, based on an independent validation set, is utilized.

It is worth mentioning that we did not use any type of contrast or intensity augmentation in the training of segmentation and regression networks, to better capture the effect of the harmonization methods and make them more sensitive to distribution shifts.

\begin{figure*}[t!]  
    \centering
    {\includegraphics[width=\linewidth]{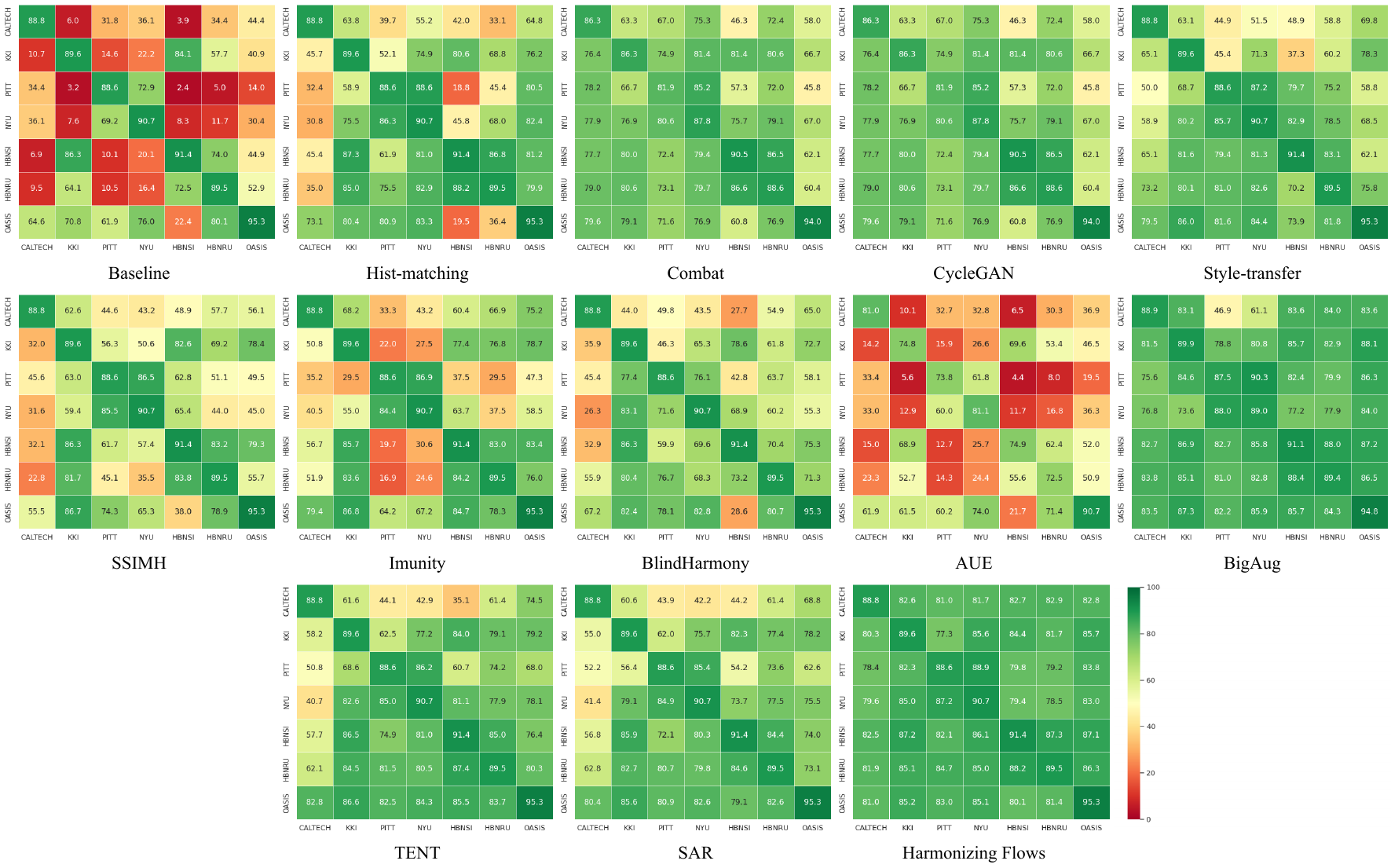}}
    \caption{\textbf{Cross-site brain MRI segmentation matrix across the compared methods.} Each cell indicates the segmentation result (DSC \%) when the source dataset (in the \textit{rows}) is used to harmonize each target dataset (in \textit{columns}).}
    \label{fig:comparing methods}
\end{figure*}

\subsection{Results}

In this section, we reported the empirical results of the experiments performed. In particular, we first resorted to the cross-site brain MRI segmentation results to evaluate the performance of different harmonization approaches, as the segmentation task is a good indicator of harmonization performance. Following the main results, we conducted a series of comprehensive ablation studies to empirically support our choices. Then, to show the task-agnostic nature of our method, we evaluated different harmonization strategies on the task of neonatal cross-site brain age estimation. Last, we took a closer look at the distance between the intensity histograms of the harmonized images with the source images. Surprisingly, closer intensity histograms between harmonized images with the source domain does not necessarily correlate with better harmonization performance or higher target task performance, such as segmentation or regression.

\subsubsection{Performance on the segmentation task}
\mypar{Main results}

To evaluate the proposed harmonization method, cross-site brain MRI segmentation performance has been obtained before and after the harmonization. As the segmentation networks are fixed, the segmentation results' improvement indicates the impact of the harmonization. Segmentation results obtained with the images harmonized by different methods are reported in Table \ref{tab:Results-comparison-adults}.


\noindent \textit{Comparison to harmonization methods.} Before looking at the segmentation results, we would like to highlight that most existing methods rely on assumptions that could limit their scalability and practicality in real-life scenarios. First, some methods must access at least one image of the source domain during the adaptation, thereby not being completely \textit{source-free} ($\gS\gF$). Also, most harmonization techniques need to access the target domains during training, while ideally, the potential target domains should remain unknown, which we refer to as \textit{unknown-domains} ($\gU\gD$). We relax all these assumptions by proposing a method that is \textit{source-free}, \textit{task-agnostic}, and unaware of potential target domains during training. 
First, from Table \ref{tab:Results-comparison-adults}, we can observe that the proposed approach improves the segmentation results by more than 45\% in terms of DSC over the baseline, i.e., without harmonization. Furthermore, it consistently outperforms other harmonization approaches by a significant margin, 
in terms of both segmentation metrics. Particularly, the average gained improvement compared to CycleGAN, the next best-performing method, is larger than 8\% in terms of DSC, and 2.4 $mm$ smaller in terms of HD95. Considering that CycleGAN requires the source domain, as well as all the target domains to adapt, the differences in performance are even more important. Furthermore, if we compare it to approaches that offer the same benefits, i.e., $(\mathcal{SF})$, $(\mathcal{TA})$ and $(\mathcal{UD})$, such as Hist matching, and BlindHarmony these differences increase up to 20\%.

\begin{table*}[h!]
\caption{\textbf{Performance overview on the cross-site neonatal MRI segmentation task.} Segmentation performance, in terms of DSC and HD95 metrics, across different harmonization approaches. To facilitate the strengths and weaknesses of different methods, we also indicate whether they are \textit{source-free} $(\mathcal{SF})$, \textit{task-agnostic} $(\mathcal{TA})$, and can handle \textit{unknown-domains} $(\mathcal{UD})$, as well as the different strategy they fall in. The best results are highlighted in \textbf{bold}.} 
\begin{center}
\footnotesize
{\begin{tabular}{l|ccc|c|c|c|c}
\toprule
 \multirow[b]{2}{*}{Method} & \multirow[b]{2}{*}{$\mathcal{SF}$} &  \multirow[b]{2}{*}{$\mathcal{TA}$} & \multirow[b]{2}{*}{$\mathcal{UD}$} & \multicolumn{2}{c|}{T1-w} & \multicolumn{2}{c}{T2-w} \\
 \cmidrule(l{3pt}r{3pt}){5-6}\cmidrule(l{3pt}r{3pt}){7-8}
 & & & &  DSC (\%) & HD95 (mm) & DSC (\%) & HD95 (mm) \\
 \midrule 

Baseline & -- & -- & -- & 34.7\ppm6.5 & 38.4\ppm15.8 & 44.2\ppm17.1 & 30.8\ppm17.0\\
Hist matching~\cite{nyul2000new} &  \cmark &  \cmark  & \cmark  & 42.9\ppm8.0 & 32.9\ppm17.6 & 84.3\ppm4.1 & 4.7\ppm2.3 \\
Combat~\cite{pomponio2020harmonization} &  \xmark &  \cmark  & \xmark  &  83.0\ppm4.5 & 2.7\ppm1.4  &  88.0\ppm1.5 & 2.0\ppm1.1 \\
Cycle-GAN~\cite{modanwal2020mri} &  \xmark &  \cmark  & \xmark  &  77.2\ppm6.5 & 4.8\ppm3.1 &  88.3\ppm1.5 & 1.8\ppm0.9 \\
Style-transfer~\cite{liu2021style} &  \xmark &  \cmark  & \xmark  &  56.7\ppm9.0 & 19.8\ppm9.9  &  88.6\ppm1.4 & 1.5\ppm0.2 \\
SSIMH$_\text{ MLMI'22}$~\cite{guan2022fast} &  \xmark &  \cmark  & \cmark  & 38.7\ppm9.7 & 30.2\ppm13.3 & 61.9\ppm8.5 & 17.9\ppm6.6 \\
Imunity$_\text{MedIA'23}$~\cite{cackowski2023imunity} & \xmark &  \cmark  & \cmark & 41.8\ppm7.9 & 23.8\ppm9.5 & 87.1\ppm3.0 & 2.43\ppm1.80 \\
BlindHarmony$_\text{ ICCV'23}$~\cite{jeong2023blindharmony} &  \cmark &  \cmark  & \cmark  & 56.1\ppm9.2 & 28.4\ppm12.6 & 88.0\ppm1.8 & 2.9\ppm1.6\\
AUE ~\cite{wang2019aleatoric}  &  \cmark &  \xmark  & \cmark & 38.0\ppm6.1 & 25.5\ppm7.0 &  68.7\ppm5.2 & 15.5\ppm1.9  \\
BigAug$_\text{ TMI'20}$~\cite{zhang2020generalizing} &  \cmark &  \xmark  & \cmark  & 84.1\ppm3.8 & 2.2\ppm53.9 & \textbf{90.0\ppm1.3} & \textbf{1.4\ppm0.3} \\
TENT$_\text{ ICLR'21}$~\cite{wang2021tent} &  \cmark &  \xmark  & \cmark  &  76.3\ppm3.8 & 4.7\ppm2.2  &  88.1\ppm1.6 & 2.4\ppm1.6 \\
SAR$_\text{ ICLR'23}$~\cite{niu2023towards} &  \cmark &  \xmark  & \cmark  & 72.3\ppm5.1 & 5.1\ppm2.2 & 88.9\ppm1.7 &  1.7\ppm0.6\\
\midrule
\textbf{Harmonizing Flows} &  \cmark &  \cmark  & \cmark  & \textbf{84.8\ppm2.3} & \textbf{2.1\ppm1.0} & 89.6\ppm1.4 & \textbf{1.4}\ppm0.4  \\

\bottomrule
\end{tabular}}
\end{center}

\label{tab:Results-comparison-neonatal-segmentation}
\end{table*}



\noindent \textit{Comparison to test-time domain adaptation and generalization approaches.} To better evaluate the method, we have compared it against common strategies to tackle distributional shift, including: domain generalization~\cite{zhang2020generalizing}, Test time Augmentation~\cite{wang2019aleatoric}, and Test time adaptation~\cite{wang2021tent, niu2023towards}. This comparison highlights the importance of image harmonization and examines whether harmonization can be replaced by any of these strategies. 
These results, which are depicted at the bottom of Table \ref{tab:Results-comparison-adults}, demonstrate that the proposed harmonization method also outperforms well-known test-time domain adaptation and generalization strategies in the task of cross-site brain segmentation. 
Furthermore, individual cross-site brain MRI segmentation results are depicted in Fig \ref{fig:comparing methods}, for a better interpretability of the per-site results. In every matrix, the diagonal elements represent the segmentation of intra-site brain MRI and establish the upper bound of segmentation results when test images originate from the source domain. Then, the elements outside the diagonal indicate the segmentation result when a given dataset (indicated in the \textit{rows}) is used to harmonize the target datasets (in \textit{columns}). As the non-diagonal elements approach the diagonal, it can be interpreted as an enhanced capacity of the method to handle distributional shifts. Based on these results, we can state that our approach proved the most effective in this aspect, and consistently across all source-domain datasets.


\mypar{Validation on a different population and image modality}

To investigate the scalability of our harmonization method, we evaluate it using different population demographics and image modalities. Particularly, 
we resort to the neonatal brain MRI segmentation task, which differs from the more traditional adult brain MRI segmentation. 
Furthermore, this evaluation has been repeated for both T1-weighted and T2-weighted MRIs to explore whether the proposed approach can yield satisfactory performance for both modalities. According to the results from this experiment, which are reported in Table \ref{tab:Results-comparison-neonatal-segmentation}, our method demonstrates comparable performance for both modalities in neonatal brain MRI segmentation. More concretely, the proposed harmonization strategy obtains the best results in T1-weighted, whereas it ranks first among compared harmonization methods and second by a small margin among all compared methods for both metrics in T2-weighted images. These values underscore the effectiveness of our proposed approach across diverse populations and modalities.

\mypar{Ablation studies}

\begin{figure}[h!]  
    \centering
    {\includegraphics[width=0.6\linewidth]{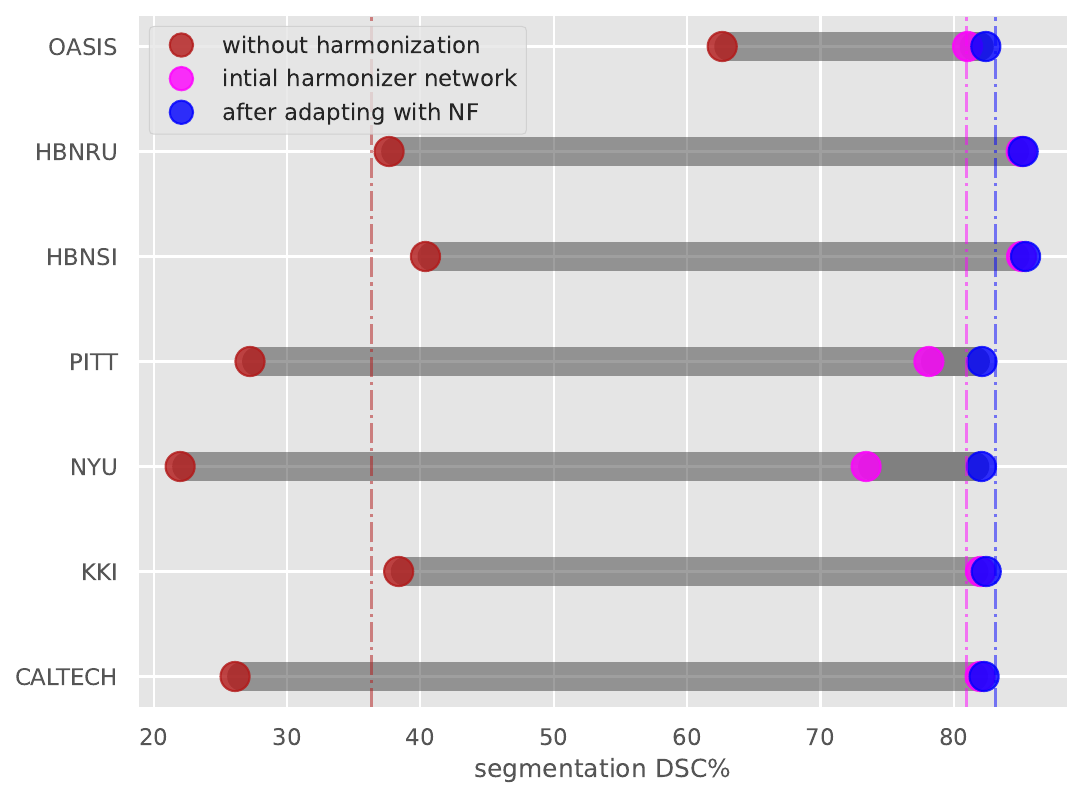}}
    \caption{\textbf{Effect of each component of the harmonizing flow.} Particularly, we depict the improvement gained using the proposed pre-trained harmonizer network ($\sim$44 DSC\%), and the adaptation using normalizing flows ($\sim$2.2 DSC\%).}
    \label{fig:normalizing flow effect}
\end{figure}

\mypar{I-Impact of normalizing flows.} This section assesses the impact of each component of the proposed method, which is achieved by comparing the cross-site brain MRI segmentation results obtained: \textit{i)} when images are not harmonized, \textit{ii)} when harmonized just with the proposed pre-trained harmonizer network $\theta^{init}$, or \textit{iii)} harmonized with the proposed method. The results of this ablation study, which are depicted in Fig. \ref{fig:normalizing flow effect}, empirically support that the proposed NF-based models can serve as an effective mechanism to adapt the harmonizer network. First, the proposed approach to pre-train the harmonizer network yields significant enhancements compared to non-harmonized images (nearly 44\% of DSC on average) while, despite its simplicity, there is no need to know the target domain in advance. Furthermore, adapting the harmonizer network using the proposed NF model further improves the cross-site MRI segmentation results, with nearly 2.2\% of DSC on average, illustrating the effectiveness of the proposed harmonization strategy.

\begin{figure}[h!]  
    \centering
    {\includegraphics[width=0.6\linewidth]{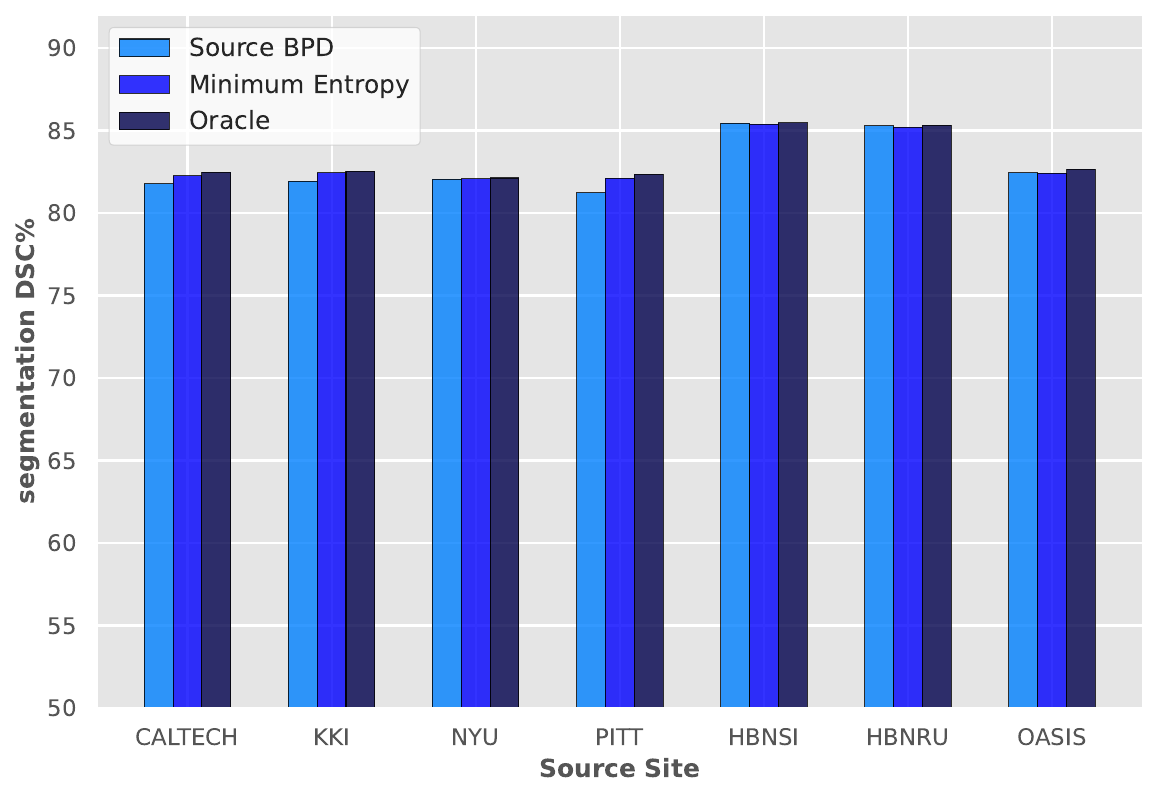}}
    \caption{\textbf{The effect of different stopping criteria to stop the harmonizer network adaptation.} Oracle represents selecting the best iteration based on the performance of the target task (i.e., segmentation in this example), which serves as the upper bound. Both criteria, \textit{source BPD} and minimum entropy, provide good stopping points, with a slight advantage of the minimum entropy criterion.}
    \label{fig:stopping-criteria}
\end{figure}

\mypar{II-Adaptation stopping criterion.} Now we explore the crucial issue of determining the appropriate iteration to stop the adaptation, specially as updating the harmonizer network is conducted in an unsupervised manner. 
Three different criteria to stop adapting the harmonizer network are explored. The first criterion is to stop in an iteration where the Shannon entropy of the target task predictions (segmentation in this case) reaches its minimum. According to~\cite{wang2021tent}, the Shannon entropy of the predictions of the target task is highly correlated with its performance. As it does not require any labeled data, it provides an \textit{a priori} reliable stopping criterion for fine-tuning the harmonizer network. 
As a second alternative, we stopped the harmonizer network adaptation when the target \textit{BPD} reaches the observed \textit{BPD} on the source domain (which can be computed during training time using a validation set). As opposed to the first criterion, this criterion is task-agnostic and well-suited for unsupervised tasks or scenarios where entropy calculations are not applicable (i.e., regression problems). Finally, we directly used the target task (segmentation) performance and stopped the adaptation once it achieved the highest DSC score which we refer to as the \textit{Oracle}. Please bear in mind that this criterion is impractical in real-world scenarios, as one may not have access to segmentation labels\footnote{Please note that using segmentation, or other kind of labels, for the stopping criteria would make the model task-dependent.}, and it only intends to serve as an upper bound to show how much we can gain using a well-defined stopping criterion. As shown in  Fig. \ref{fig:stopping-criteria}, despite minimum entropy being a slightly better criterion compared to \textit{source BPD}, both yield similar performances, with the \textit{source BPD} providing a more general strategy, as it is not tailored to a task that requires probabilistic outputs. In summary, both stopping criteria prove to be viable options, as their outcomes closely resemble those of the \textit{Oracle}. 

\begin{figure}[h!]  
    \centering
    {\includegraphics[width=0.6\linewidth]{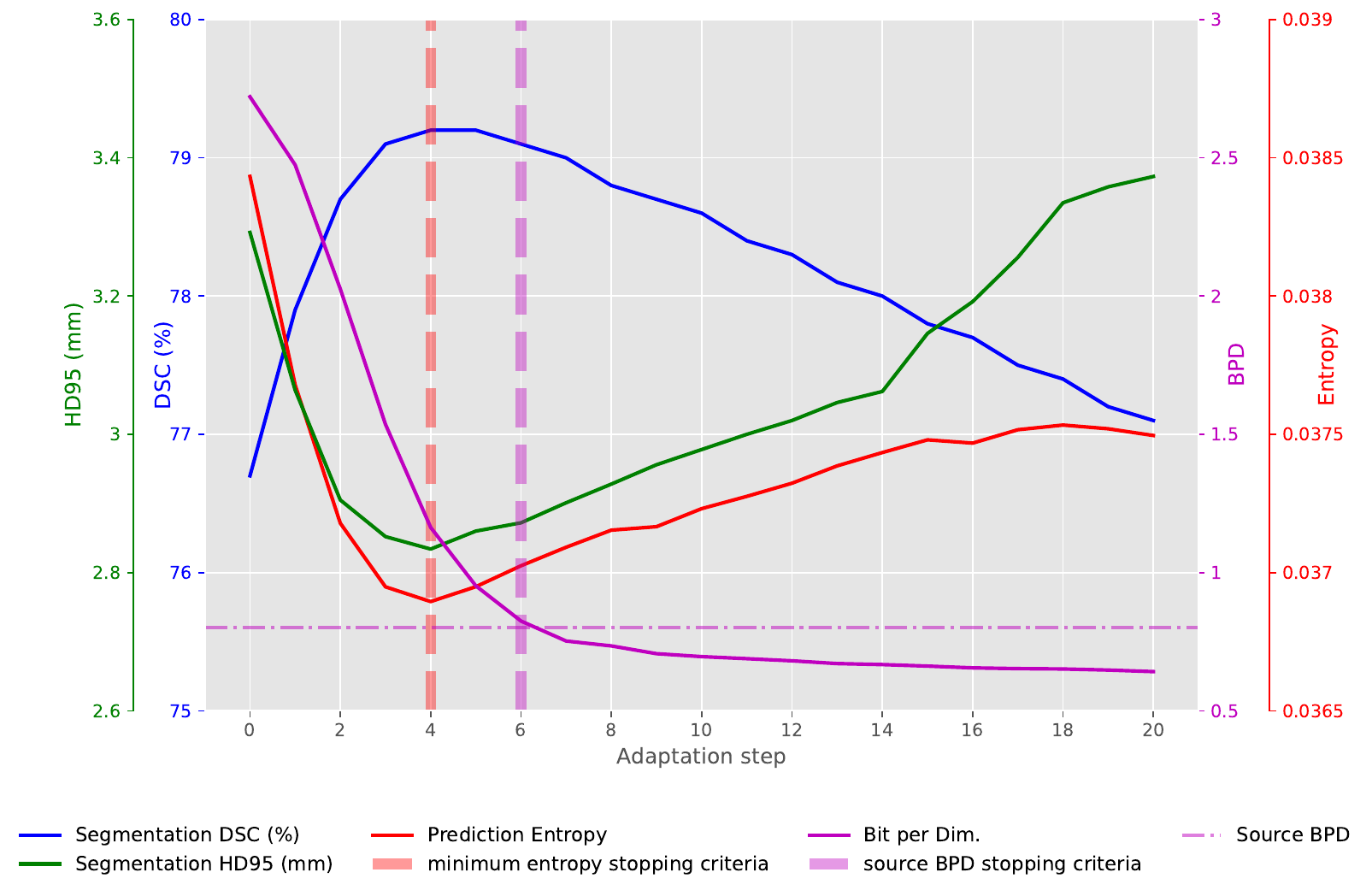}}
    \caption{\textbf{Which is the best metric as stopping criteria?} This plot depicts different metrics during the adaptation of the harmonizer network (from HBNSI to NYU). Step zero corresponds to using the initial harmonizer network without adaptation. The vertical lines show the stopping time-points based on two proposed stopping criteria: minimum entropy of the predictions (\textit{red})and reaching source BPD \textit{purple}). }
    \label{fig:step-sadaptation-metrics}
\end{figure}

Furthermore, in Fig. \ref{fig:step-sadaptation-metrics} we show, for a given scenario --harmonizing HBNSI dataset to NYU-- the strong correlation between the segmentation metrics with both the segmentation prediction entropy and proximity to the source BPD. In particular, the evolution of the segmentation metrics is depicted in green (HD95) and blue (DSC) curves, with the values of the entropy of the segmentation predictions in red, and the target BPD in purple. We can observe that the point for which these metrics are optimal (i.e., minimum entropy and target BPD matching source BPD), is actually close. Nevertheless, we advocate that, for more general use, looking at the source BPD should be a preferred option, as it does not depend on the target task.


\mypar{III-Ablation studies on components and hyper-parameters.} This section aims to empirically support the choices made in the proposed harmonization strategy.

\begin{figure}[h!]
  \centering
  \begin{subfigure}[b]{0.24\linewidth}
    \includegraphics[width=\linewidth]{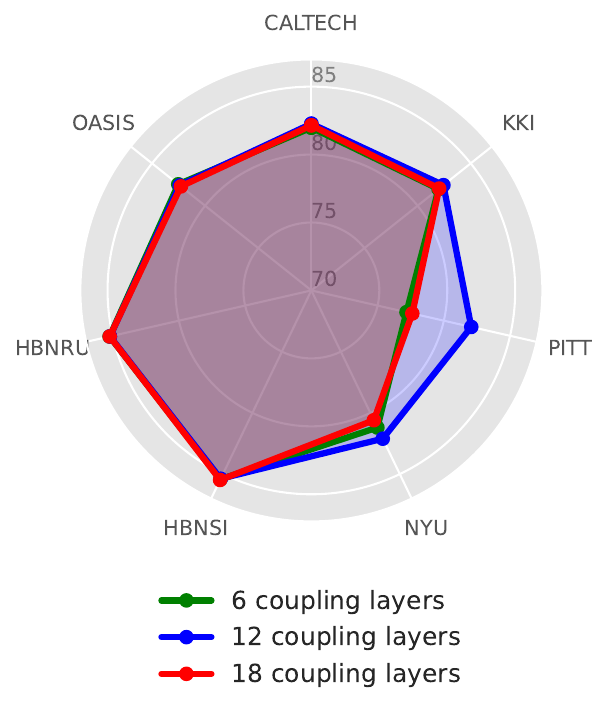}
    \caption{Number of coupling layers in normalizing flow network}
    \label{fig:components-flow_layers}
  \end{subfigure}
  \begin{subfigure}[b]{0.24\linewidth}
    \includegraphics[width=\linewidth]{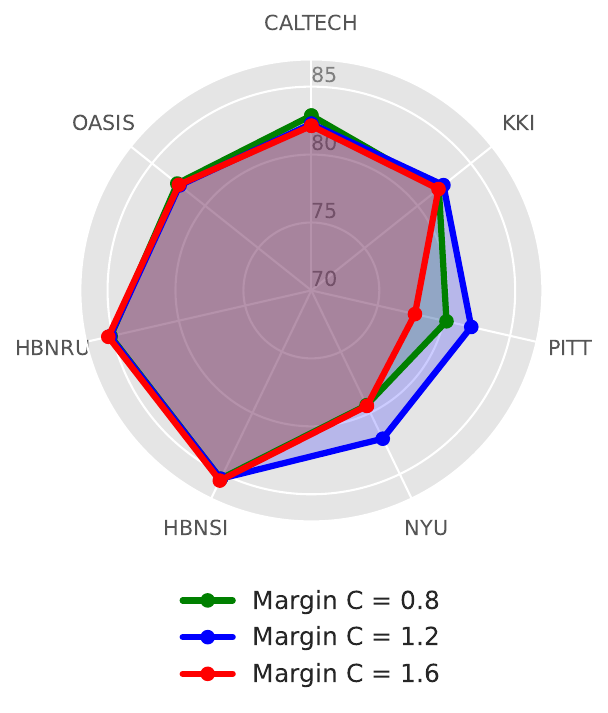}
    \caption{Margin $\mathrm{c}$ for guiding the normalizing flow network}
    \label{fig:hyper-params-margin_c}
  \end{subfigure}
  \begin{subfigure}[b]{0.24\linewidth}
    \includegraphics[width=\linewidth]{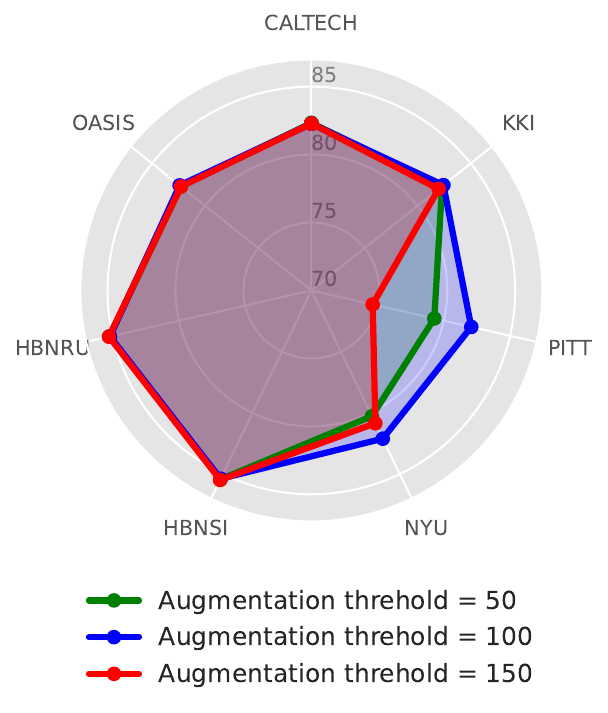}
    \caption{Augmentation threshold to consider out of distribution}
    \label{fig:hyper-params-augmentation_threhold}
  \end{subfigure}
  \begin{subfigure}[b]{0.24\linewidth}
    \includegraphics[width=\linewidth]{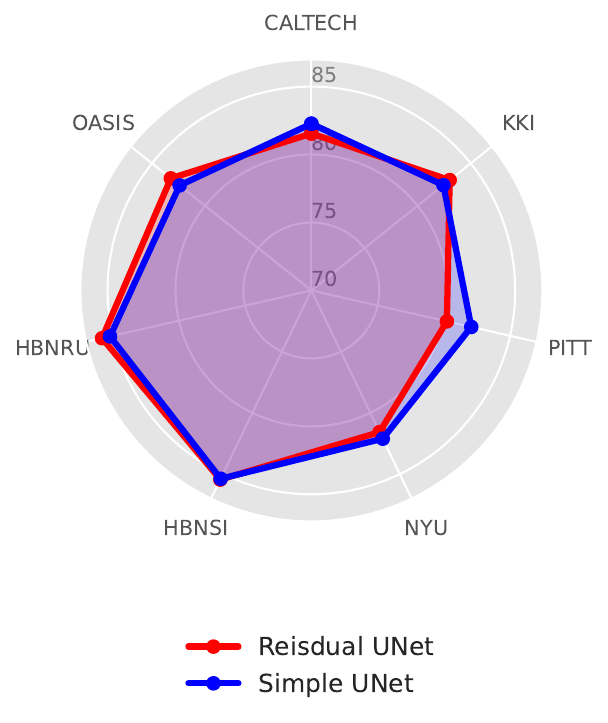}
    \caption{Architecture of the harmonizer network}
    \label{fig:components-harmonizer_network}
  \end{subfigure}
  
  \caption{\textbf{Ablation study on architecture components and hyper-parameters of the harmonizing flow.} The best performance is obtained when the number of coupling layers in the normalizing flow is set to 12, the guiding margin $c$ to 1.2, the augmentation threshold is 100, and a simple UNet is chosen as the harmonizer network.}
  \label{fig:components-and-hyper-params}
\end{figure}

\noindent\textbf{III.a-Flow Depth:}. We explored the effect of the number of coupling layers in the normalizing flow network in learning the distribution of the source domain. Particularly, we investigated using 6, 12, and 18 coupling layers in the normalizing flow network and its effect on the distribution learning capacity of the model and harmonization process. As can be seen in Fig. \ref{fig:components-flow_layers}, 12 coupling layers were the optimal choice and resulted in better harmonization. We believe that using 6 coupling layers does not fully capture the source domain distribution. On the other hand, using 18 coupling layers makes the training process harder, and it probably needs more data to be fully trained. 

\noindent\textbf{III.b-Margin $\mathbf{c}$ :}
In this section, we investigate the influence of the margin $\mathrm{c}$ on the guidance of the normalizing flow process. Our initial choice of 1.2 for $\mathrm{c}$ at the first level stemmed from observing an average source BPD of approximately 0.8 when the normalizing flow was trained without guidance. Consequently, we opted for a value of 1.2. To further investigate the effect of this value, we conducted additional experiments with margin values $\mathrm{c}$ set to 0.8 and 1.6. As illustrated in Fig. \ref{fig:hyper-params-margin_c}, the initial choice yielded better performance in comparison to 0.8 and 1.6.

\begin{figure*}[h!]
\centering
    {\includegraphics[width=.99\linewidth]{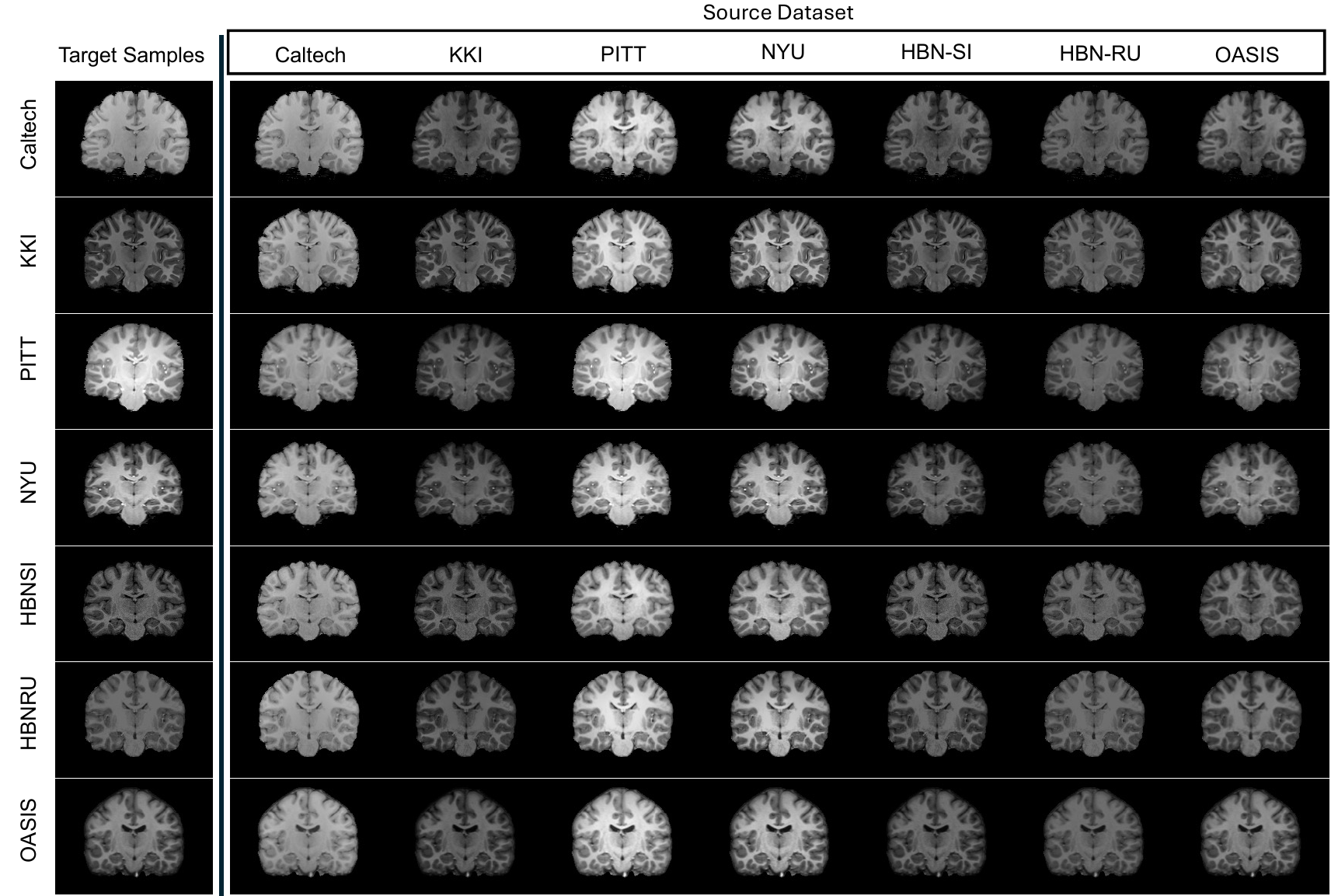}}
    \caption{This figure showcases examples of harmonized images generated by the proposed method. The first column presents the sample images from various target domains. The subsequent columns display the images harmonized to specific source domains, as indicated at the top of each column. Therefore, each row maintains consistent anatomical structures, while each column shares the same visual characteristics.}
    \label{fig:qualitative-results-adults}
\end{figure*}

\noindent\textbf{III.c-Augmentation threshold:} Defining how much augmentation is enough for an augmented image to be considered as out of distribution to train the normalizing flow is an important step. 
This threshold, which is defined in terms of mean squared distance with original images, was first selected as 100, by visual inspection of the images.  
After the initial choice, we explored how adjusting this hyper-parameter affects the constraint on the normalizing flow network and the harmonization process. We conducted experiments using augmentation thresholds of 50 and 150, whose results (Fig. \ref{fig:hyper-params-augmentation_threhold}) demonstrate 100 to be an optimal choice for this hyperparameter. 

\noindent\textbf{III.d-Harmonizer Network:} We investigated two U-shaped architectures for the harmonizer network. First, a conventional UNet architecture was employed. Subsequently, like the conference version of this paper~\cite{beizaee2023harmonizing}, we utilized a modified UNet to extract two separate sets of values. The final layer of the network ($\beta$) serves as a bias value, matching the input image's dimensions. Additionally, a scalar value $\alpha$, derived from the network's middle layer, acts as a scale parameter. In this way, the harmonizer's output can be expressed as $h_{\theta}(\vx) = \alpha * \vx + \beta$. As depicted in Fig. \ref{fig:components-harmonizer_network}, both options are effective, with a slight superiority of the simple UNet. We believe that this marginal superiority is due to the greater degrees of freedom in simple UNet, which might be better for transferring complex distributions. 

\mypar{Qualitative results.} 

\begin{figure}[h!]
    \centering
    {\includegraphics[width=0.6\linewidth]{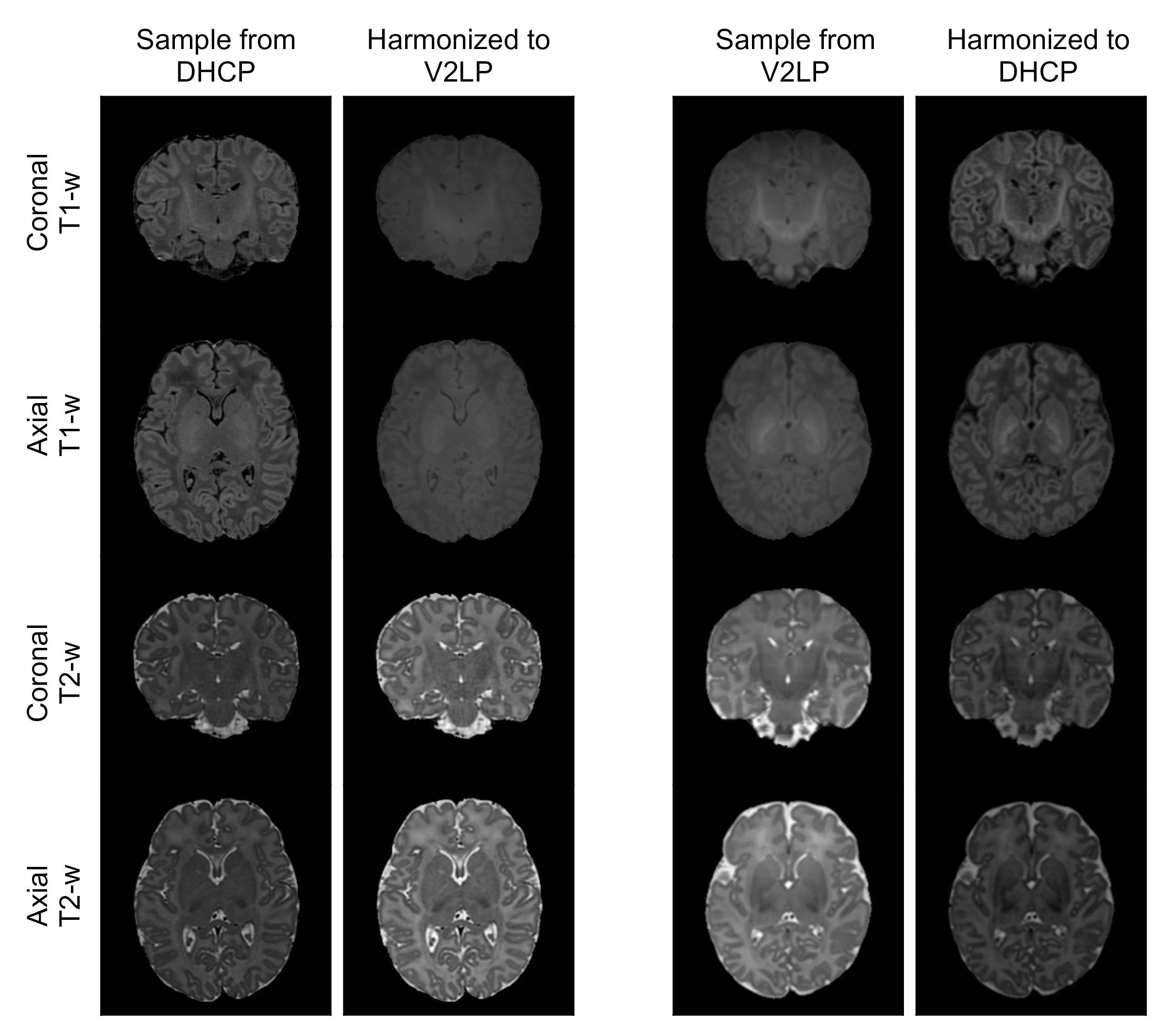}}
    \caption{Visual examples of neonatal MRI brain images harmonized using the proposed method. The left section displays the coronal and axial planes for both modalities of a sample image from the DHCP dataset (target) alongside their harmonized counterparts to the V2LP dataset (source). The right section shows the reverse, with V2LP images harmonized to the DHCP dataset.}
    \label{fig:qualitative-results-neonates}
\end{figure}

Fig. \ref{fig:qualitative-results-adults} showcases the instances of harmonized images using the proposed method across different source and target sites, for the adult brain MRIs. In particular, we randomly picked a sample from each site and then mapped it to different target sites. As can be seen, the harmonized samples in each column share the same visual characteristics, while on each row, the details of the harmonized images are preserved. 
These qualitative results illustrate that, regardless of the source or the target domains, the proposed method consistently produces reliable harmonized images, which is supported quantitatively by the comprehensive empirical validation conducted. 
Furthermore, visual examples depicted in Fig. \ref{fig:qualitative-results-neonates} (neonatal brains) demonstrate the effectiveness of our approach in harmonizing inter-site images, regardless of the modality and plane used.

In \autoref{fig:harmonized_sample_for_methods}, we have visualized a harmonized sample from the target domain (CALTECH here) to the source domain (KKI here) using different harmonization methods. The harmonized image using our proposed method appears to have the closest visual characteristics with the source domain compared to other harmonization methods. Combat\cite{pomponio2020harmonization} is excluded from this figure as it attempts to remove variations between domains and cannot transform the target domain to match the source domain.


\begin{table*}[h!]
\caption{\textbf{Performance overview on the cross-site neonatal age estimation task.} Brain age estimation performance, in terms of Mean Absolute Error (MAE) and Mean Squared Error (MSE) metrics, across different harmonization approaches and modalities. To facilitate the strengths and weaknesses of different methods, we also indicate whether they are \textit{source-free} $(\mathcal{SF})$, \textit{task-agnostic} $(\mathcal{TA})$, and can handle \textit{unknown-domains} $(\mathcal{UD})$, as well as the different strategy they fall in. The best results are highlighted in \textbf{bold}.} 
\begin{center}
\footnotesize
{\begin{tabular}{l|ccc|c|c|c|c}
\toprule
 \multirow[b]{2}{*}{Method} & \multirow[b]{2}{*}{$\mathcal{SF}$} &  \multirow[b]{2}{*}{$\mathcal{TA}$} & \multirow[b]{2}{*}{$\mathcal{UD}$} & \multicolumn{2}{c|}{T1-w} & \multicolumn{2}{c}{T2-w} \\
 \cmidrule(l{3pt}r{3pt}){5-6}\cmidrule(l{3pt}r{3pt}){7-8}
  & & & &   MAE & MSE & MAE & MSE \\
 \midrule 

Baseline & -- & -- & -- & 1.15\ppm0.91 & 2.23 & 1.67\ppm1.15 & 4.67\\
Hist. matching~\cite{nyul2000new} &  \cmark &  \cmark  & \cmark  & 1.04\ppm0.78 & 1.77 & 1.09\ppm0.84 & 2.05  \\
Combat~\cite{pomponio2020harmonization} &  \xmark &  \cmark  & \xmark  &  1.33\ppm0.84 & 2.49  &  1.28\ppm0.82 &  2.33  \\
Cycle-GAN ~\cite{modanwal2020mri} &  \xmark &  \cmark  & \xmark  & 1.05\ppm0.71 & 1.62 & 0.89\ppm0.63  &  1.24 \\
Style-transfer~\cite{liu2021style} &  \xmark &  \cmark  & \xmark   &  1.75\ppm1.17 & 5.15  &  0.96\ppm0.67 & 1.41  \\
SSIMH~\cite{guan2022fast} & \xmark &  \cmark  & \cmark  & 1.18\ppm0.96 & 2.33 & 1.41\ppm0.85 & 2.35\\
Imunity~\cite{cackowski2023imunity} & \xmark &  \cmark  & \cmark & 1.10\ppm0.81 & 1.89 & 1.55\ppm1.00 & 3.82 \\
BlindHarmony~\cite{jeong2023blindharmony} &  \cmark &  \cmark  & \cmark  & 1.19\ppm0.98  & 2.45 & 1.29\ppm0.90 & 2.62 \\
AUE ~\cite{wang2019aleatoric} &  \cmark &  \xmark  & \cmark  & 1.32\ppm1.01 & 2.91 &  1.4\ppm1.01 & 3.51 \\
BigAug~\cite{zhang2020generalizing} &  \cmark &  \xmark  & \cmark  & 1.05\ppm0.62 & 1.55 & 1.09\ppm0.80 & 1.88 \\
\midrule
\textbf{Harmonizing Flows} &  \cmark &  \cmark  & \cmark  & \textbf{1.01\ppm0.69} & \textbf{1.51} & \textbf{0.82\ppm0.68} &  \textbf{1.21} \\

\bottomrule
\end{tabular}}
\end{center}

\label{tab:Results-comparison-neonatal-age-estimation}
\end{table*}

\subsubsection{Performance on neonatal brain age estimation}

In the previous section, we demonstrated the generalization capabilities of the proposed approach, by showing its superiority when evaluating the harmonized images in adults and neonatal brain MRI segmentation, and across multiple modalities. In this section, we will take one step further and we will evaluate the quality of the harmonization based on a different task, i.e., neonatal brain age estimation, which necessitates regression neural networks and the utilization of different 2D slices (specifically the axial view). As depicted in Table \ref{tab:Results-comparison-neonatal-age-estimation}, our method also outperforms all compared methods for both metrics and both modalities, highlighting its effectiveness for various tasks and applications, as well as on different 2D planes. Thus, based on the results of the segmentation and regression tasks, we can state that the proposed harmonization strategy leads to a significantly more flexible solution with substantial improvement gains.

\subsubsection{A closer look at the harmonization performance}

In previous sections, we have assessed the quality of the harmonization based on the performance of different target tasks, i.e., adult and neonatal brain segmentation and neonatal brain age estimation. A natural question that arises is whether the harmonization performance can be quantitatively evaluated without requiring further labeled target tasks or \textit{traveling subjects}. Indeed, recent literature \cite{parida2024quantitative} in evaluating harmonization techniques has proposed the use of  the Wasserstein distance, to measure the similarity between harmonized and source image intensity histograms. The reasoning behind using this kind of divergence is two-fold. First, standard harmonization metrics, such as mean absolute error, mean squared error, or peak signal-to-noise ratio, can provide high-quality measurements, but at the price of requiring paired harmonized data, i.e., \textit{traveling subjects}. Second, while other metrics, such as structural similarity (SSIM), can help mitigate the need for paired data, they primarily compare structures at a higher level, potentially overlooking smaller artifacts or hallucinations introduced by generative models. In this section, we follow up on the recent work in \cite{parida2024quantitative}, and assess the harmonization performance of different strategies based on the Wasserstein distance between intensity histograms.




\begin{figure}[h!]
\centering
    {\includegraphics[width=0.6\linewidth]{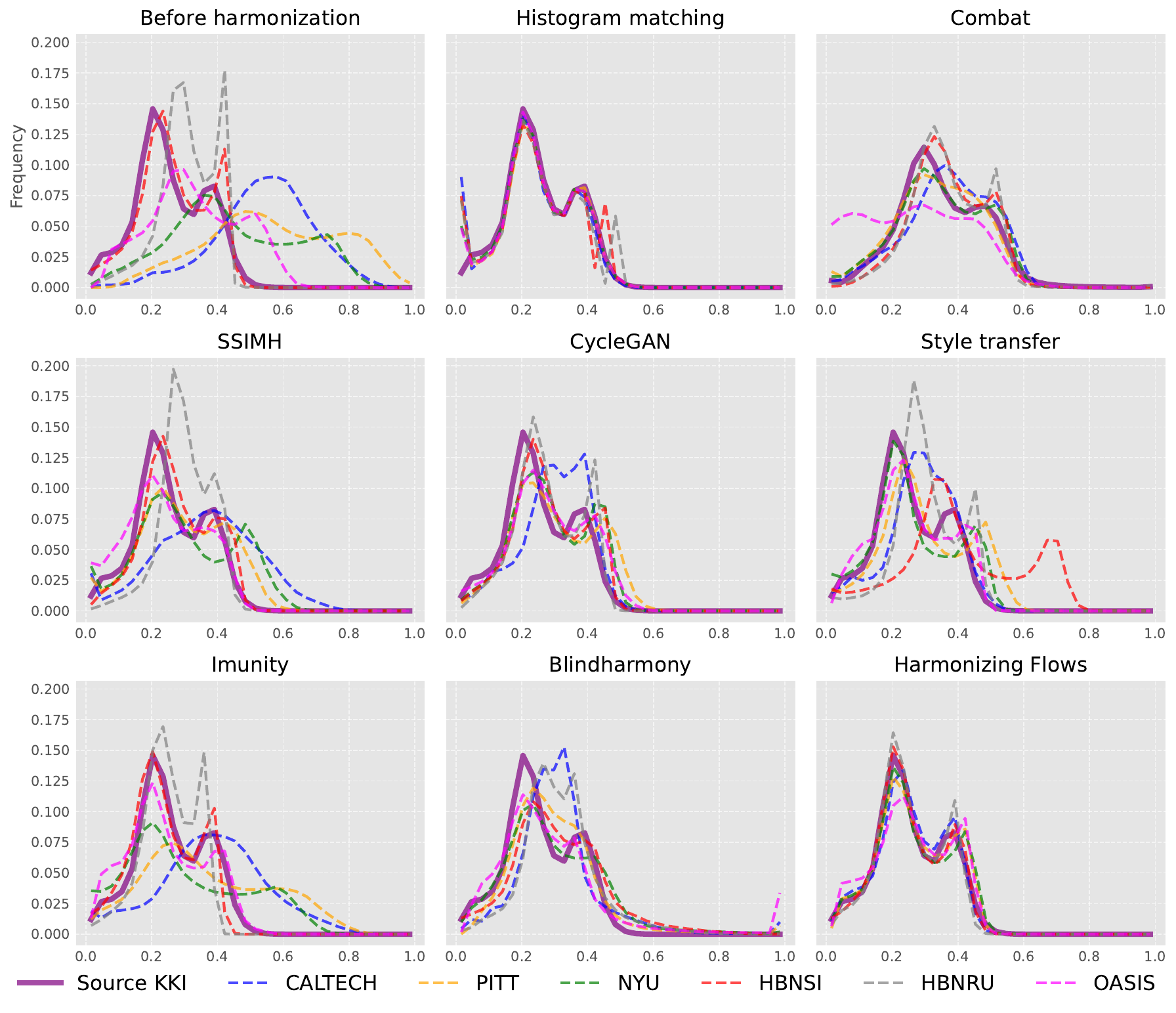}}
    \caption{Histograms of the harmonized MRIs from multiple target datasets compared to the histogram of the source MRIs (KKI dataset, \textit{in purple}) for all harmonization methods. }
    \label{fig:histograms-adults}
\end{figure}

\begin{figure*}[h!]
\centering
    {\includegraphics[width=\linewidth]{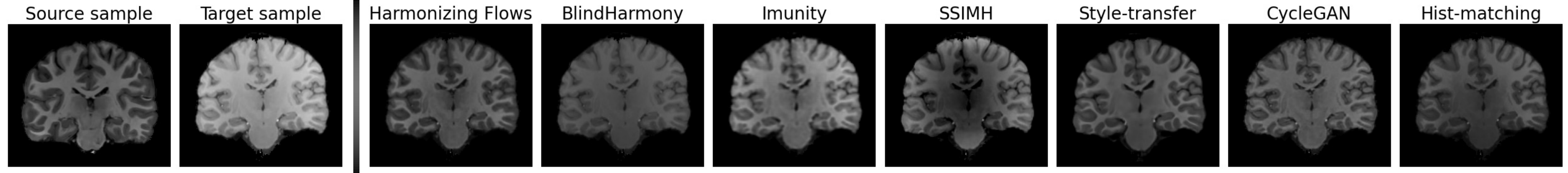}}
    \caption{Example of harmonized images using different methods. The first image is the source domain sample (KKI here), the second one shows a sample from the target domain (CALTECH here), and the rest, shows the harmonized target sample using different methods.}
    \label{fig:harmonized_sample_for_methods}
\end{figure*}

\begin{figure*}[h!]
    \centering
        \centering
        \includegraphics[width=\textwidth]{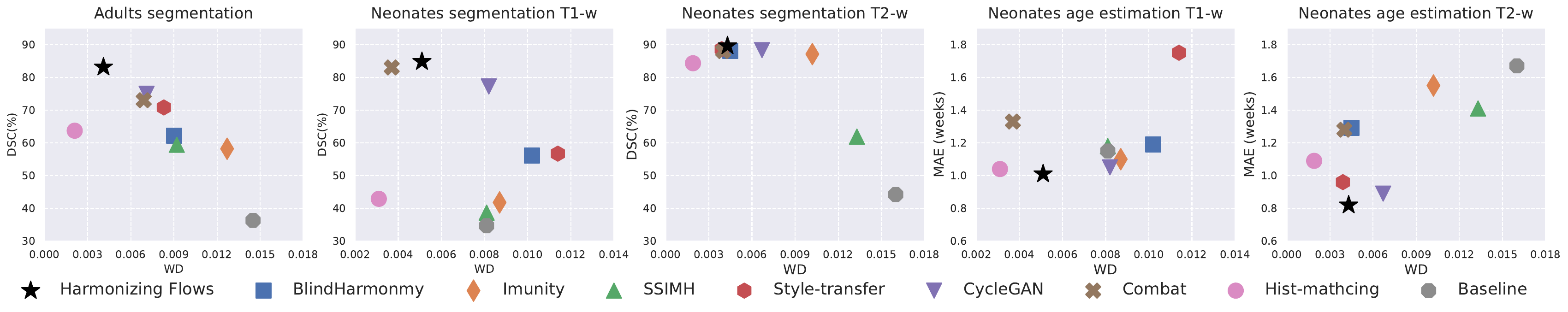}
    \caption{Segmentation (DSC\%) and Age estimation (MAE) results vs. WD of intensity histograms for compared Harmonization methods.}
    \label{fig:DSC_VS_WD-adults}
\end{figure*}

To this end, we first depict in Fig. \ref{fig:histograms-adults} the histogram distributions of the harmonized MRIs from multiple target domains compared to the MRIs from the source site (the KKI dataset in this example), across all harmonization methods. Looking at these plots, we can observe that Histogram Matching, as well as our approach, yield to the closest intensity histogram distributions to the source domain (\textit{in purple}), after harmonization. Furthermore, this behavior is consistent for all the target sites for both Histogram Matching and the proposed approach. Thus, from a pure harmonization standpoint, these plots tell us that Histogram Matching and our harmonization strategy are capable of mapping images from a target to a source site such that the harmonized intensity histograms almost match perfectly those of the source domain.  

Nevertheless, perfect alignment between intensity histograms is not the ultimate goal. The true value of the harmonization lies in its ability to enhance the performance of subsequent tasks, such as segmentation, classification, or brain age estimation in our context, that rely on harmonized images when dealing with multi-centric data. However, if we analyze the performance on target tasks when employing harmonized data, we can easily observe that histogram matching indeed generates harmonized images that lead to poor performance on multiple target tasks. For example, in adult brain segmentation, resorting to histogram matching as a harmonization method results in an average DSC of 63.6, substantially lower than other methods which, a priori, had the highest misalignment between intensity histograms of the harmonized images with the source one in terms of WD (e.g, SSIMH or Imunity). 

To better understand whether a correlation between the intensity histogram distance and the target task performance exists, we plot \textit{task-performance vs. intensity-histogram-distance (WD)} in Figure \ref{fig:DSC_VS_WD-adults}. Indeed, we can observe from this figure that, in most cases, a closer intensity histogram (i.e., lower WD) corresponds to better performance in the target task (i.e., higher DSC in segmentation and lower MAE in brain age estimation). A method warranting further discussion is histogram matching, which shows a weak correlation, if any, between intensity histogram distance in terms of WD and target task performance. 
Its good performance in the WD metric is somehow expected, as it directly optimizes the histogram of intensities for MRI harmonization. Nevertheless, we believe that solely forcing the histograms to be close is not necessarily a good condition to yield usable harmonized images, as the spatial variation across intensities is not considered when matching distributions. Harmonized images depicted in \autoref{fig:harmonized_sample_for_methods} could further illustrate this point visually. The remaining harmonization strategies, however, seem to be generally correlated, with approaches obtaining lower WD values also achieving better performances on the respective target tasks. In particular, our proposed approach always ranks among the top-three approaches in terms of WD, yielding the best performance in segmentation and age estimation tasks among compared harmonization methods.
These findings are based on the metrics proposed in the recent work \cite{parida2024quantitative}, and may not hold for a more in-depth list of harmonization metrics. Ideally, while some other metrics could have been explored to evaluate the harmonization performance, some of the metrics used depend on \textit{traveling-subjects}, which limits their applicability to many scenarios. We also stress that having an in-depth evaluation of the harmonization performance and assessing the best harmonization metric, is not within the scope of this work. In contrast, we believe that based on the results reported in this work, solely relying on the closeness of intensity histograms may not be sufficient, and evaluating the quality of the harmonization on target task indicators (e.g., DSC in segmentation or MAE in regression tasks) could serve as a more general manner to assess the harmonization quality.

\begin{figure}[h]
\centering
    {\includegraphics[width=0.6\linewidth]{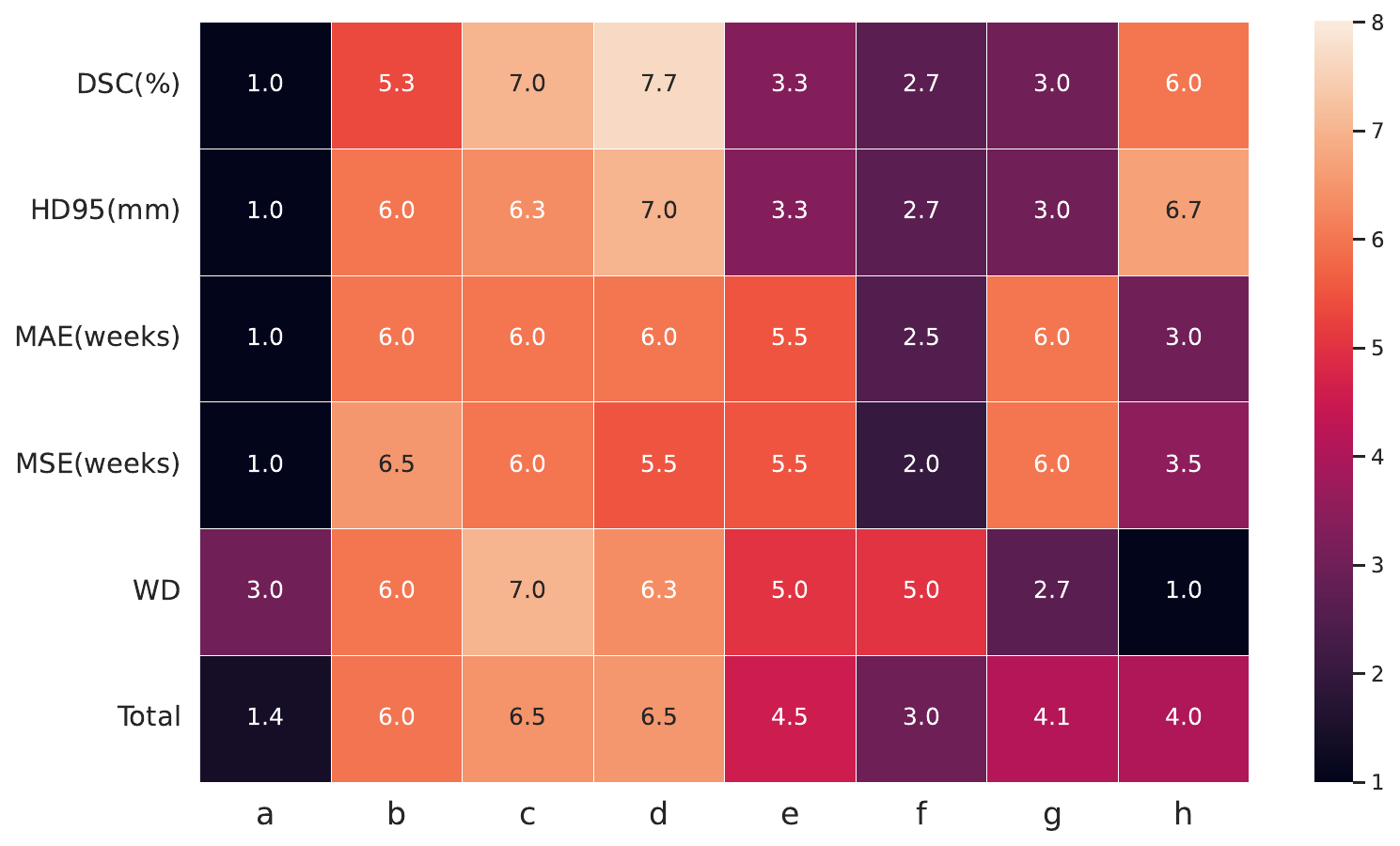}}
    \caption{Friedman Rank for the compared harmonization methods: a) Harmonizing Flows (Ours), b) BlindHarmony, c) Imunity, d) SSIMH, e) Style-transfer, f) CycleGAN, g) Combat, h) Hist-matching.}
    \label{fig:Friedman-Rank}
\end{figure}

\subsubsection{Friedman Ranking}

To fairly compare the performance of the different harmonization methods across various metrics and tasks, we resort to the Friedman Rank \cite{friedman1937use,friedman1940comparison}, which has been employed for this purpose in the literature \cite{wang2022usb,murugesan2024neighbor}. 
The \textbf{Friedman Rank} 
is defined as:

\[ \text{rank} = \frac{1}{S_m} \sum \text{rank}_i \]

where \( S_m \) is the number of evaluation settings and \(\text{rank}_i\) is the rank of a method in the \(i\)-th setting. Thus, the lower the rank obtained by an approach, the better the method is. In our scenario, we have 13 different settings: DSC and HD95 in 3 segmentation tasks, MAE and MSE in 2 regression problems, and 3 WD values in harmonization (one for adults and two for neonatal brains).  
The results from the Friedman Ranking across all analyzed methods are depicted in Figure \ref{fig:Friedman-Rank}. As it can clearly observed, our proposed method (\textit{last column}) achieves the best Friedman rank among all compared harmonization methods, demonstrating their overall superiority across different scenarios.

\section{Conclusion}

In this work, we proposed a novel harmonization method that leverages Normalizing Flows to guide the adaptation of a harmonizer network. Our approach is source-free, task-agnostic, and works with unseen domains. These characteristics make our model applicable in real-life problems where  the source domain is not accessible during adaptation, target domains are unknown at training time and harmonization is task-independent. Furthermore, another advantage of our method over the existing approaches is that it only requires images from one source domain, and one modality, at the training time. 

 Through extensive comparisons with other harmonization methods, as well as test-time domain adaptation and generalization approaches, our method consistently proved its superiority in multiple medical image problems, yet relaxing the strong assumptions made by existing harmonization strategies. 
 Furthermore, we validated the scalability of our proposed method by evaluating it on a different population (neonates), different modalities (T1-weighted and T2-weighted MRIs), and different tasks (neonatal brain age estimation and segmentation). The results demonstrated comparable performance across diverse populations and modalities, highlighting the robustness and versatility of our approach. Qualitative results further supported the reliability and effectiveness of our method, illustrating consistent and reliable image-to-image mappings across different target domains. Last, even compared to recent test-time adaptation strategies, empirical results suggest that the proposed method is a powerful alternative to deal with the presence of domain drifts, more particularly for MRI multi-site harmonization. 
 

In conclusion, our proposed harmonization method offers a promising solution for addressing distributional shifts in medical image analysis, paving the way for improved performance and generalizability across diverse datasets, and enabling the use of large-scale multi-centric studies.

\section*{Acknowledgements}
This work is supported by the National Science and Engineering Research Council of Canada (NSERC), via its Discovery Grant program and IVADO - Grants for Fundamental Research Projects. We also thank Calcul Quebec and Compute Canada.

\bibliographystyle{unsrt}  
\bibliography{main}

\end{document}